\documentclass{elsarticle}
\usepackage{times}
\usepackage{soul}
\usepackage{url}
\usepackage[utf8]{inputenc}
\usepackage[small]{caption}
\usepackage{graphicx}
\usepackage{amsmath}
\usepackage{amsthm}
\usepackage{booktabs}
\usepackage{algorithm}
\usepackage{algorithmic}
\usepackage[switch]{lineno}

\usepackage{algorithm}
\usepackage{algorithmic}
\usepackage{comment}
\usepackage{hyperref}
\hypersetup{colorlinks=true}

\usepackage{newfloat}
\usepackage{listings}
\usepackage{todonotes}
\usepackage{reledmac}
\usepackage{paralist}
\usepackage[utf8]{inputenc} %
\usepackage[T1]{fontenc}    %
\usepackage{booktabs}       %
\usepackage{amsfonts}       %
\usepackage{nicefrac}       %
\usepackage{microtype}      %

\usepackage{booktabs} 
\usepackage{subcaption}
\usepackage{graphicx} %
\usepackage{pgfplots} %
\pgfplotscreateplotcyclelist{rw}
{solid, mark repeat = 1, mark phase = 1, mark = *, black\\
	solid, mark repeat = 1, mark phase = 1, mark = square*, black\\}

\usepackage{tikz} %
\usetikzlibrary{shapes,arrows}
\usepackage{pgfplots}
\usepackage{bussproofs}
\usepackage{xspace}
\usepackage{amsmath}
\usepackage{amssymb}
\usepackage{mathtools}

\usepackage{color}

\usetikzlibrary{arrows,shapes,calc}
\usetikzlibrary{trees,positioning,fit}
\tikzstyle{arrow}=[draw,-to,thick]
\tikzstyle{embedding} = [draw, minimum width=8mm, minimum height=6mm]
\tikzstyle{nnop} = [draw, minimum width=8mm, minimum height=8mm, rounded 
corners]
\tikzstyle{block} =
[rectangle, draw,
minimum width=9mm, text centered, rounded corners, minimum height=2.0em]
\tikzstyle{smallblock} =
[rectangle, draw,
minimum width=9mm, text centered, minimum height=2.0em]

\tikzstyle{line}=[draw]
\tikzstyle{cloud} =
[draw, text centered, ellipse, minimum height=2.0em, minimum width=9em]

\usepackage[]{listings}

\newcommand{\mloop}{\mathit{loop}}
\newcommand{\mloopt}{\mathit{loop2}}
\newcommand{\mcompr}{\mathit{compr}}
\newcommand{\mcond}{\mathit{cond}}
\newcommand{\mdiv}{\mathit{div}}
\newcommand{\mmod}{\mathit{mod}}
\newcommand{\mathleft}{\@fleqntrue\@mathmargin0pt}

\newenvironment{mypar}[2]
  {\begin{list}{}%
    {\setlength\leftmargin{4mm}
    \setlength\rightmargin{4mm}}
    \item[]}
  {\end{list}}

\usepackage{setspace}

\definecolor{Code}{rgb}{0,0,0}
\definecolor{Decorators}{rgb}{0.5,0.5,0.5}
\definecolor{Numbers}{rgb}{0.5,0,0}
\definecolor{MatchingBrackets}{rgb}{0.25,0.5,0.5}
\definecolor{Keywords}{rgb}{0,0,0}
\definecolor{self}{rgb}{0,0,0}
\definecolor{Strings}{rgb}{0,0.63,0}
\definecolor{Comments}{rgb}{0,0.63,1}
\definecolor{Backquotes}{rgb}{0,0,0}
\definecolor{Classname}{rgb}{0,0,0}
\definecolor{FunctionName}{rgb}{0,0,0}
\definecolor{Operators}{rgb}{0,0,0}
\definecolor{Background}{rgb}{1,1,1}

\lstdefinelanguage{Python}{
	numbers=none,
	numberstyle=\footnotesize,
	numbersep=1em,
	xleftmargin=1em,
	framextopmargin=20em,
	framexbottommargin=20em,
	showspaces=false,
	showtabs=false,
	showstringspaces=false,
	frame=o,
	tabsize=4,
	basicstyle=\ttfamily\small\setstretch{1},
	backgroundcolor=\color{Background},
	commentstyle=\color{Comments}\slshape,
	stringstyle=\color{Strings},
	morecomment=[s][\color{Strings}]{"""}{"""},
	morecomment=[s][\color{Strings}]{'''}{'''},
	morekeywords={import,from,class,def,for,while,if,is,in,elif,else,not,and,or,print,break,continue,return,True,False,None,access,as,,del,except,exec,finally,global,import,lambda,pass,print,raise,try,assert},
	keywordstyle={\color{Keywords}\bfseries},
	morekeywords={[2]@invariant,pylab,numpy,np,scipy},
	keywordstyle={[2]\color{Decorators}\slshape},
	emph={self},
	emphstyle={\color{self}\slshape},
}
\journal{International Journal of Approximate Reasoning}

\begin{document}

\begin{frontmatter}

\title{Alien Coding}
\author[label1]{Thibault Gauthier}
\author[label2]{Miroslav Ol\v{s}\'ak}
\author[label1]{Josef Urban}

\affiliation[label1]{organization={Czech Technical University in Prague},%
	country={Czech Republic}}

\affiliation[label2]{organization={Institut des Hautes Etudes Scientifiques},%
	city={Paris},
	country={France}}

\begin{abstract}
	We introduce a self-learning algorithm for synthesizing programs that provide explanations for
	OEIS sequences. The algorithm starts from scratch initially generating programs at random. Then it runs many iterations of a
	self-learning loop that interleaves (i) training neural machine
	translation to learn the correspondence between sequences and the
	programs discovered so far, and (ii) proposing many new programs for
	each OEIS sequence by the trained neural machine translator. The
	algorithm discovers on its own programs for more than 78000 OEIS
	sequences, sometimes developing unusual programming methods. We
	analyze its behavior and the invented programs in several
	experiments.
\end{abstract}

\end{frontmatter}

\section{Introduction}

\begin{mypar}{10mm}{5mm}
{\it
  Galileo once said, "Mathematics is the language of
Science." Hence, facing the same laws of the physical world, alien mathematics
must have a good deal of similarity to ours.

{\hfill\rm--  R. Hamming - Mathematics on a Distant Planet~\cite{hamming1998mathematics}}
}
\end{mypar}

\begin{mypar}{10mm}{5mm}
{\it
Certainly, let us learn proving, but also let us learn guessing.

{\hfill\rm--  G. Polya - Mathematics and Plausible Reasoning~\cite{polya1990mathematics}}
}
\end{mypar}

Proposing explanations for abstract patterns is one of the main
occupations of mathematicians. Polya famously argued that educated guessing is
as important skill as theorem proving~\cite{polya1990mathematics}.

Integer sequences are
perhaps the most common kind of mathematical patterns.  In this work,
we describe a self-learning system -- a positive feedback loop between
learning, search and verification -- that can in 190
iterations discover from scratch programs and explanations for more than 78000
sequences from the On-Line Encyclopedia of Integer Sequences
(OEIS)~\cite{oeis}.
Such positive feedback loops between learning, search and verification
are, in our opinion, among the most interesting objects of study in
today's Artificial Intelligence (AI). In automated theorem proving and
reasoning, such loops -- typically interleaving deductive search with
learning its guidance -- have been researched for over 15 years in
several contexts~\cite{Urban07,US+08,KaliszykUMO18,JakubuvU19}, often leading to large self-improvements of
the theorem proving systems. However, guessing and conjecturing~\cite{ReynoldsBNBT19,DBLP:conf/mkm/UrbanJ20,abs-2212-11151,abs-2210-03590} are
still relatively unexplored skills and subsystems in today's mainstream automated
theorem provers (ATPs). 

Today, perhaps the most widely known and cited examples of self improvement which combine learning and search 
are in games such as Go and Chess, which were proposed as an
interesting abstract playground for developing machine intelligence
already by Turing~\cite{turing1950computing}.
An early example of such a \emph{reinforcement learning} (RL) system is 
\textsf{MENACE}~\cite{10.1093/comjnl/6.3.232} which 
can learn to play noughts and crosses on its own and much faster than brute-forcing all 
possible solutions. Recently, with a residual network as a machine 
learner, \textsf{AlphaGoZero}~\cite{silver2017mastering} has learned to play Go 
better than professional players using only self-play. In this process, the 
system discovered many 
effective moves that went against 3000 years of human wisdom. The early 3-3 
invasion was considered a bad %
opening move but is now frequently used by professionals.
Similarly, our system presented here is capable of discovering \emph{on its own} possibly unusual (\emph{alien}) mathematical formulas and
programs without any human guidance, inventing, for example, over 20 different definitions of (pseudo-)primes (Table~\ref{tab:primes1}).

Ultimately, RL systems still suffer from some human biases, not least
because their different search and learning modules were coded by
humans. Inspired by biological evolution, ``AI
life''~\cite{langton1997artificial} attempts to remove almost all
human bias. It only sets the rules (physics) of the simulated world
and lets the programs evolve freely. A famous example of such a system
is Conway's Game of Life~\cite{conway1970game}. Since such systems are
open-ended, they yet have to be turned into useful tools. Indeed, it
is even a hard problem to detect when such systems invent useful
programs.

In this work, we use the OEIS setting to experiment with two different
high-level heuristics that guide the overall evolution of the system
that invents explanations: \emph{Occam's razor} (the \emph{law of
  parsimony}), i.e., the length of the explanations, and
\emph{efficiency}, i.e., the time which it takes to evaluate an
invented explanation as a program that takes integers as inputs.
Occam's razor
has been one of the most important heuristics guiding scientific research in general.
In a machine learning setting, this can be seen as a form of
regularization.  A mathematical proof of this principle relying on
Bayesian reasoning and assumptions about the computability of our
universe is provided in Solomonoff's theory of inductive
inference~\cite{SOLOMONOFF19641}.
There is however an interesting
tradeoff between short explanations and their efficiency when used as
declarative programs and building blocks of more complex programs. A
short but inefficient program may be too expensive to evaluate,
preventing us from verifying whether it really is an explanation for a
longer integer sequence, and thus further learning from such an
explanation.  In this work, we therefore use a combined fitness
function for our growing and evolving population of programs: a
program is the fittest if it is either the shortest one or the most
efficient one.

\paragraph{Overview and Contributions}

\begin{figure}[t]
\centering
\begin{tikzpicture}[scale=0.99,every node/.style={scale=0.99},node 
distance=1cm]
\node [draw] (1){Search};
\node [right of=1, node distance=2cm] (2) {};
\node [draw,right of=2, node distance=2cm] (3) {Check};
\node [draw,below of=2] (4) {Learn};
\draw[-to,thick] (1) to node[yshift=3mm] {programs} (3);
\draw[-to,thick] (3) to node[xshift=8mm,yshift=-2mm] {examples} (4);
\draw[-to,thick] (4) to node[xshift=-6mm,yshift=-2mm] {weights} (1); 
\end{tikzpicture}
\caption{The three phases of the self-learning loop. \label{fig:overview}}
\end{figure}
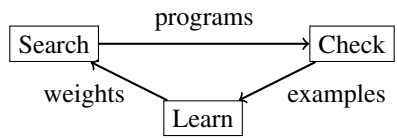

Our approach relies on a self-learning loop  to learn how to find programs 
generating integer sequences. It alternates between
the three phases represented in Figure~\ref{fig:overview}. During
the \emph{search phase}, our machine learning model synthesizes
programs for integer sequences. In this work, we predominantly use
\emph{neural machine translation} (NMT) as the machine learning
component.
For each OEIS sequence, the NMT trained on previous examples typically creates 240
candidate programs using \emph{beam search}. In the first iteration
of this loop, programs are randomly constructed. Then, during the
\emph{checking phase}, the proposed millions of programs are checked to see if 
they generate their target sequence or any other OEIS sequences.  The
smallest and fastest programs generating an OEIS sequence are kept to
produce the training examples. In the \emph{learning phase}, NMT trains on these examples %
to translate the ``solved'' OEIS sequences into
the best program(s) generating them. 
This updates the weights of the NMT network
which influences the next search phase. Each iteration of the
self-learning loop leads to the discovery of more solutions, as well as
to the optimization of the existing solutions.

This work builds mostly on our previous work in~\cite{abs-2202-11908}. Our
contributions are the following.
The tree neural network is replaced by a relatively 
fast encoder-decoder NMT network 
(bidirectional LSTM with attention) and the previously used MCTS search is 
replaced by a
relatively wide beam search during the NMT decoding.
Our objective function lets us collect both the smallest and
fastest programs for each sequence instead of only the smallest programs 
in~\cite{abs-2202-11908}.
We introduce local and global definitions to the programming language.
These definitions are created automatically and the newly defined
symbols (tokens) can be used by the NMT network to build more complex programs.
Finally, our longest run finds solutions for more than 78000 OEIS sequences in 
190 iterations, and all our experiments have so far together produced  
solutions for 84587 OEIS sequences. This is more than three times the number 
(27987) invented in our first experiments~\cite{abs-2202-11908}.

The rest of this paper is structured as follows. Section~\ref{sec:components} 
describes in more detail the basic components of
our system and our self-learning loop for discovery of mathematical explanations:
the OEIS datasets, our formula (programming) language %
including %
local and global definitions, and the
checking phase which implements the language of our expressions and their 
efficient evaluation.
In Section~\ref{sec:nmttask}, we describe the NMT network, its parameters, and 
the different training techniques such as continuous training and combined 
training. These methods contribute significantly to the final performance of 
the NMT runs.
In Section~\ref{sec:tnn}, we 
do side experiments with the original tree neural network~\cite{abs-2202-11908}, %
optimizing its 
different parameters such as the embedding size, %
and the search 
strategy. The final experiment of this section relies on the best parameters and 
is run for 500 iterations (instead of 25 in~\cite{abs-2202-11908}) to create a 
strong baseline.
In Section~\ref{sec:exp-nmt}, we give details about our long-running 
experiments with the NMT network. 
We observe how they improve over the TNN baseline and analyze how
introducing local macros and global macros affects the performance.
And last, in Section~\ref{sec:analysis}, we analyze the solutions produced 
during these runs and their evolution over time.
We also measure the generalization performance of the smallest
and the fastest solutions and show that the smallest solutions generalize better.

\section{Components}\label{sec:components}
In this section, we give a technical description of the components of
our system: the OEIS datasets, the programming language and its 
representations, and the
checking phase.

\subsection{The OEIS Dataset}
The OEIS is a repository created and maintained by Neil Sloane
where amateur and professional mathematicians can contribute integer
sequences.  There are currently more than 350,000 sequences in this
repository.  Each entry contains the terms of the sequence and a short
English description.  It is referenced by an A-number. For example, A40
is the reference for the sequence of prime numbers. Additional information may 
be
provided such as: alternative descriptions, links to related OEIS entries, links
to papers where the sequence was investigated and in about one-third
of the cases a program for generating the sequence is provided. These
programs are written in many different languages such as \textsf{PARI}, 
\textsf{Matlab}, \textsf{Haskell}, \ldots . In our experiments, we do not train 
on any human-written programs. Thus, our problem set consists of all
OEIS integer sequences (351663 as of March 2022). Solutions/programs for 
sequences are discovered automatically and incrementally by our system starting 
from scratch.

\subsection{The Programming Language and its Python and NMT Representations}\label{sec:program}

A formal description of the programming language used in this paper is
given in \cite{abs-2202-11908}. It is minimalistic by design, to
avoid human-informed bias.  Since many examples given in this paper
require an understanding of the language, we briefly summarize it
here. The language contains two variables x and y, that
can take as values arbitrary-precision integers. It includes the
standard operators $0,1,2,+,\times,\mmod,\mdiv$ (integer division) and the
conditional operators
$\mcond (a,b,c) :=\ \mbox{if } a \leq 0 \mbox{ then } b \mbox{ else }
c$. These programming operators follow the standard semantics of most
programming languages (including C and Python).  In this programming
language, an expression $p$ can either be evaluated to an integer if
given specific values for $x$ and $y$ or can be used to create a binary
function $f$ defined by $f(x,y) = p$. The three looping operators in this
language have functional arguments designated by $f,g$ and
integer arguments noted $a,b,c$. Looping expressions may themselves be
used as arguments of looping operators allowing for arbitrary
nesting of loops.

	\paragraph{The $\mloop$ Operator}
	This operator takes three arguments: one function and two integers.
	\begin{align*}
	\mloop (f,a,b) := &\ b &\mbox{if } a \leq 0\\
	&\ f (\mloop(f,a-1,b),a) &\mbox{ otherwise}
	\end{align*}	
This definition is almost the same as the one used to define primitive 
recursion in the standard theory of primitive recursive functions.
For more clarity and portability, we can translate this
    construction to Python. We capitalize the variables in $a$ and
    $b$, which play a different role than the variables in $f$, to
    avoid undesirable variable capture.  Python's
    implementation $F$ of the function that can be derived from
    the expression $\mloop(f,a,b)$ is as follows:

\begin{lstlisting}[language=Python]
def F(X,Y) = 
    x = b[x/X,y/Y]
    for y in range (1,a[x/X,y/Y] + 1)
        x = f(x,y)
    return x
\end{lstlisting}
Some common uses of
$\mloop$ include: $2^x$ written as $\mloop (2\times x, x, 1)$
and $x!$ written as $\mloop (y\times x, x, 1)$.

\paragraph{The $\mloop2$ Operator}
This operator takes five arguments: two functions and three integers.	
\begin{align*}
\mloopt (f,g,a,b,c) := &\ b &\mbox{ if } a \leq 0\\
&\mloopt (f,g,a-1,f(b,c),g(b,c)) &\mbox{ otherwise}
\end{align*}
This operator starts with the pair of numbers $(b,c)$ and
updates their values $a$ times using the functions $f$ and $g$
before returning $b$.  This is a generalization of the $\mloop$
operator. Given $g$ such that $g(x,y) = y+1$, we have
$\mloop (f,a,b) = \mloopt (f,g,a,b,1)$.  Therefore, 
the $\mloop$ operator could be removed from the language
without affecting its expressiveness. It is kept in the
language as it is a natural and useful instantiation of
$\mloopt$.  Python's implementation $F$ of the function
derived from the expression $\mloopt (f,g,a,b,c)$ is:
\begin{lstlisting}[language=Python]
def F(X,Y) = 
    x = b[x/X,y/Y]
    y = c[x/X,y/Y]
    for _ in range (1,a[x/X,y/Y] + 1)
        x = f(x,y)
        y = g(x,y)
    return x
\end{lstlisting}
        The following constructions have a natural implementation
        using $\mloopt$.  They are however difficult to express
        using $\mloop$ and would generally require 
        encodings such as the Cantor pairing function:  The Fibonacci
        function $\mathit{Fibonacci}(x)$ can be implemented by the
        program $\mloopt (x + y, x, x, 0, 1)$, and the power
        function $x^y$ by the program
        $\mloopt (x \times y, y, y, 1, x)$.

\paragraph{The $\mcompr$ Operator}
The comprehension operator takes two arguments: one function and one integer.
\begin{align*}	
\mcompr (f,a) :=\ &\mathit{failure} &\mbox{if } a < 0\\
&\mathit{min} \lbrace m\ |\ m \geq 0 \wedge f(m,0) 
\leq 0 \rbrace 
&\mbox{if } a = 0\\
&\mathit{min} \lbrace m\ |\ m > \mcompr (f,a-1) \wedge f(m,0) 
\leq 0 \rbrace&\mbox{otherwise}
\end{align*}
The comprehension expression finds the ${a+1}^{th}$ smallest
nonnegative integer $m$ satisfying the predicate $f(m,0)$. If the
value of $a$ is $0$ then this behaves like the minimization operator
$\mu$ in the theory of general recursive functions, thus making the
language Turing-complete. It gives a natural way of constructing
increasing sequences of numbers (i.e., sets) from a predicate.  In particular, suppose
we have constructed the function
$f_{prime}(x,y) = \mbox {if } x \mbox{ is prime then } 0 \mbox{ else } 1$. Then the
expression $compr(f_{prime},x)$ constructs the sequence of primes
as the value of $x$ increases from 0 to infinity. %
Note that the operator thus behaves similarly %
to the set comprehension operator in set theory.  The Python's
implementation $F$ of the function derived from the expression
$\mcompr (f,a)$ is:
\begin{lstlisting}[language=Python]
def F(X,Y):
    x,i = 0,0
    while i <= a[x/X,y/Y]:
        if f(x,0) <= 0:
            i = i + 1
        x = x + 1
    return x - 1
\end{lstlisting}
\paragraph{Linear Representation of Programs for NMT}

We use prefix notation (with argument order reversed) to represent a program as a
sequence of tokens. The main advantage of this approach is that the
notation does not require the use of parentheses. For example, the prefix
notation for the program $\mloop (x \times y, x, 1)$ is
$\mloop\ 1\ x\, \times\, y\ x$.  When using NMT, the 14
operators/actions/tokens
$[0,1,2,+,-,\times,\mdiv,\mmod,\mcond,\mloop,x,y,\mcompr,\mloopt]$
are represented by capital letters from A to N. Thus, the program for
factorial is written as J B K F L K.

\paragraph{Definitions}
To allow code reuse, human %
programmers introduce %
definitions.
We allow %
NMT to produce %
definitions in two different settings and experiments: one using \emph{local
  definitions} and the other using \emph{global definitions}.  We have
decided that such definitions will be arbitrary sequences of actions. 
Throughout the paper, the more specific term \emph{macros} may be used
to refer to such definitions. 
This is quite a powerful setup since such macros can represent
not just subprograms, but also subprograms with holes.
A program can be always constructed by a sequence of actions, but not every 
sequence of actions
is a program.

\begin{figure}
    \centering
    \includegraphics[width=0.5\textwidth]{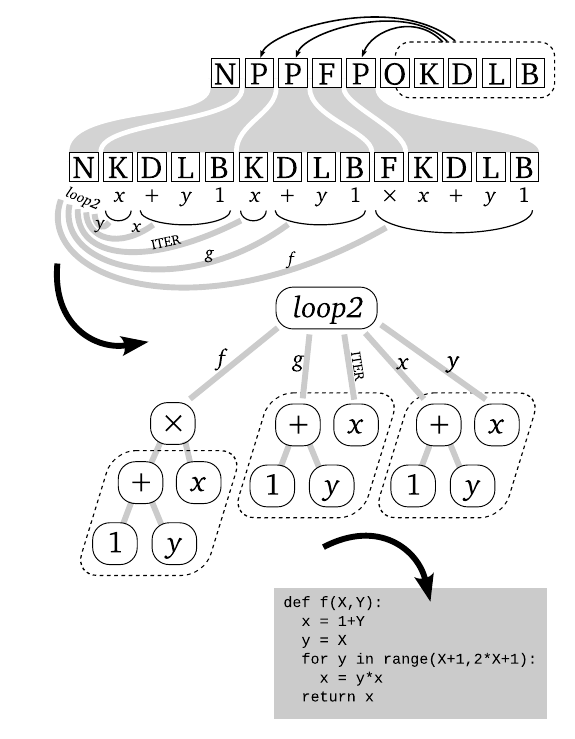}
    \caption{Representing local macros. A macro version and expanded
      version of a program invented for A1813 ($a(n) =
      (2n)!/n!$). Note that the macro here (K D L B) is not a proper
      program, only a sequence of actions.}
    \label{fig:exlmacros}
\end{figure}

\paragraph{Local Definitions}
In this setting, we add to the programming language ten tokens
representing ten possible local definitions (macros) that may include
preceding macros. A special action/token is used as a separator
between the different macros.  The macros in the generated programs
are unfolded before the checking phase takes place. Note that the
naming of such macros is often inconsistent across different
programs,\footnoteA{E.g., in program $P_1$, a macro called $m$ could
  be used to define $n!$, while in $P_2$, $m$ could be used to define
  $2^n$.} possibly making them harder to learn across many
examples. Figure~\ref{fig:exlmacros} shows an example of the solution
found for the sequence A1813\footnoteA{\url{https://oeis.org/A1813} -
  Quadruple factorial numbers: $a(n) = (2n)!/n!$.} which involved the
synthesis of a local macro that is then used three times in the body
of the invented program.

\begin{figure}
    \centering
    \includegraphics[width=0.7\textwidth]{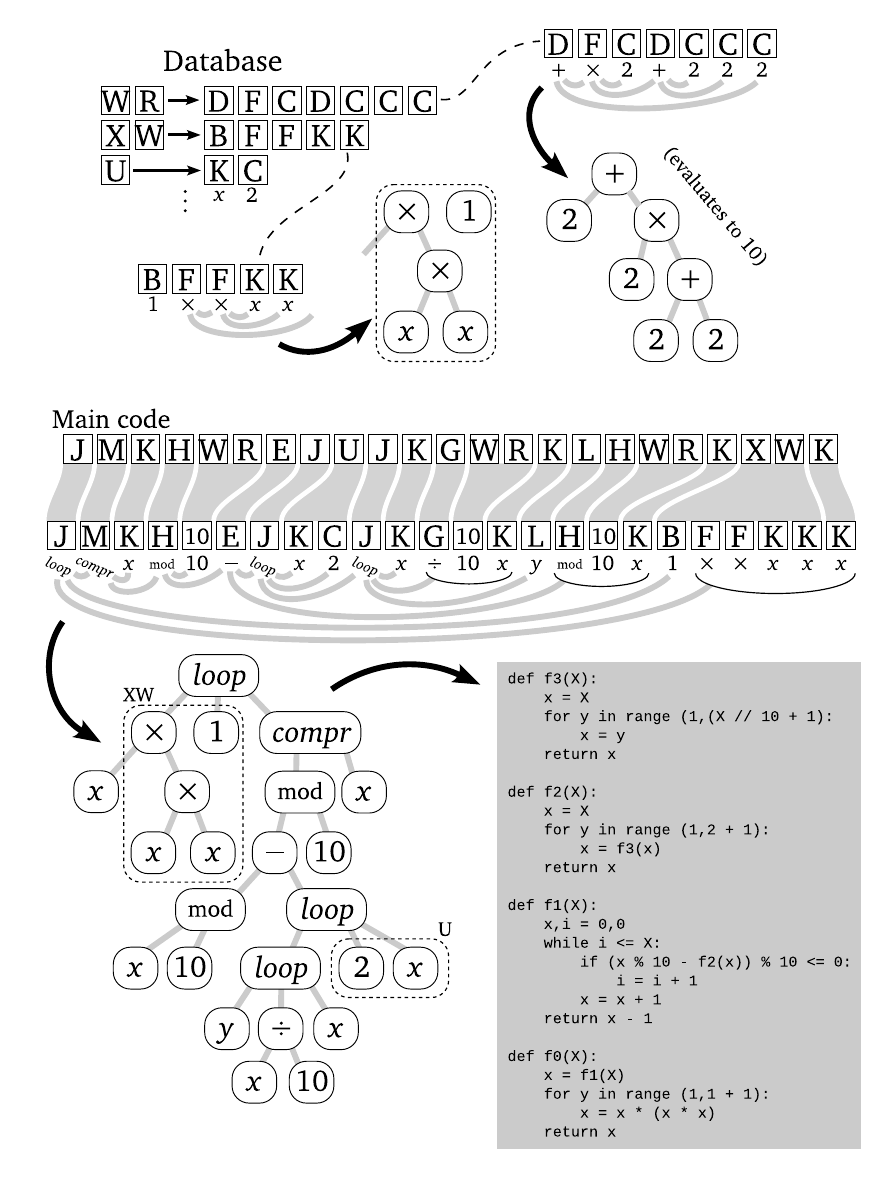}
    \caption{Representing global macros. A macro version and expanded
      version of a program invented for A14187 (cubes of palindromes). Note that two macros here (K C and B F F K K) are not proper
      programs, while the third one (D F C D C C C) is a program that evaluates to 10.}
    \label{fig:exgmacros}
\end{figure}

\paragraph{Global Definitions}
In this setup, we allow for arbitrarily many macros stored in a global
array and shared across all programs. This makes the naming of the
macros consistent in all programs, possibly making them easier to
learn. %
Programs may refer to any macro stored in the global array, 
by writing its index in base 10. %
This again requires 10 
additional actions (one for each digit) and a special action to separate 
the references to the macros.  As in the local definition setup, the
global macros may contain references to macros with lower indices.
Figure~\ref{fig:exgmacros} shows an example of the solution
found for the sequence A14187\footnoteA{\url{https://oeis.org/A14187} -
  Cubes of palindromes.} which involved
three global macros that are altogether used five times in the body
of the invented program.

Introduction and use of local macros for a particular input integer
sequence is completely a ``local'' decision of the trained NMT that
generates the particular program. In the global case, we however need
more coordination to introduce the global macros consistently. This
can be done in various ways and we now use the following method. At
every iteration of the overall loop, we add the ten most frequent
sequences of actions to the array of global macros.
To force the network to learn to use the global macros, we greedily
(starting from the macros with the lowest indices) recognize sub-sequences
of actions that correspond to the macros, and replace them with the
macros' names (indices) in the programs.

\subsection{The Program Checker}

In its most basic form the checker takes a program and a sequence and
checks if that program generates the sequence. Since programs may
depend on two variables, we say that the program $f(x,y)=p$ generates the finite
sequence %
$(s_x)_{0\leq x\leq n}$ if and only if
\[\forall x\in \mathbb{Z}.\ 0 \leq x \leq n \Rightarrow f(x,0) = s_x\]
We say that a sequence $s$ has a solution if we have found a program $p$ generating 
it. The number of OEIS sequences with at least one solution is the number 
reported in all our experiments under the label \emph{solutions}.

\paragraph{Timeout}
The first issue when implementing a program checker is to determine
the time limit for running the (generally non-terminating) programs.
In particular, to adapt the time limit for longer sequences, we
compute the generated terms in the order $f(0,0), f(1,0), f(2,0), ...$
and stop the program if it has not generated the $n$-th term in less
than $n \times t_{\mathit{call}}$ abstract time units. This effectively means 
that we give a
timeout of $t_{\mathit{call}}$ time units per call with the time unused during
previous calls added to the timeout of the current call.

This \emph{abstract time unit} is computed to be an approximation of the
number of CPU instructions needed to perform each operation. The cost
of an operation is 1 for the $+, -, \times $ operations, it is 5 for the $ 
\mdiv, \mmod$ operations, and it is
the number of bits of the result if the result is larger than 64 bits.
Using the abstract time is also important to get accurate and
repeatable measurements of the speed of the programs.

\paragraph{Hindsight Experience Replay}
To augment the training data using a limited form of
hindsight experience replay~\cite{andrychowicz2017hindsight}, we check our program against all
OEIS sequences at the same time. This can be done effectively by
organizing the sequences into a tree of sequences
(Fig.~\ref{fig:seqtree}) and stopping the checking as soon as the
generated sequence reaches a leaf in that tree or takes a non-existing
branch in the tree. All sequences (typically at most one) found along
the path taken by the generated sequence are said to \emph{have a solution}.

\begin{figure}[t]
	\centering
	\Large
	\begin{tikzpicture}[scale=0.8,every node/.style={scale=0.8},node 
	distance=1cm]
	\node [] (1){};
	\node [below of=1] (2) {};
	\node [right of=2, node distance=3cm] (2r) {};
	\node [below of=2] (3) {};
	\node [right of=3, node distance=3cm] (3r) {};
	\node [below of=3r] (3rd) {};
	\node [below of=3rd] (3rdd) {A40};
	\node [below of=3] (3d) {};
	\node [below of=3d] (3dd) {A5843};
	\node [left of=3, node distance=3cm] (3l) {};
	\node [below of=3l] (3ld) {};
	\node [below of=3ld] (3ldd) {A45};

	\draw[-to,thick] (1) to node[xshift=3mm] {0} (2); 
	\draw[-to,thick] (1) to node[xshift=3mm,yshift=2mm] {2} (2r); 
	\draw[-to,thick] (2) to node[xshift=3mm] {2} (3);
	\draw[-to,thick] (3) to node[xshift=3mm] {4} (3d);
	\draw[-to,dotted] (3d) to node[xshift=3mm] {6} (3dd);
	\draw[-to,thick] (2) to node[xshift=-3mm,yshift=2mm] {1} (3l);
	\draw[-to,thick] (3l) to node[xshift=3mm] {1} (3ld);
	\draw[-to,dotted] (3ld) to node[xshift=3mm] {2} (3ldd);
	\draw[-to,thick] (2r) to node[xshift=3mm] {3} (3r);
	\draw[-to,thick] (3r) to node[xshift=3mm] {5} (3rd);
	\draw[-to,dotted] (3rd) to node[xshift=3mm] {7} (3rdd);
	
	\end{tikzpicture}
	\normalsize
	\caption{\label{fig:seqtree} Tree of OEIS sequences with branches
          for primes (A40), even numbers (A5843) and
          Fibonacci numbers (A45).}
\end{figure}

\paragraph{Objectives}
After each iteration we keep only the fastest and the smallest
program (which could be the same) for each sequence $s$. The
\emph{speed} of a program for $s$ is the total number of abstract time
units necessary to generate $s$. The \emph{size} of a program is the
number of operators/tokens in its linear representation.  As soon as the
checker has found a program that is a solution for a particular OEIS
sequence, we compare it with the existing solutions for that
sequence. We use the abstract time to select the fastest program among
the ones that match the solutions.  The fastest and smallest programs
are also used as training examples for the next iteration of the
loop.

\subsection{Comprehension Limit}
Evaluating each term of the sequence $\mcompr(f,0),
\mcompr(f,1), \ldots, \mcompr(f,n-1)$ separately is most of the time
too slow. This computation can be sped up using the fact that each term can be 
computed from the preceding term in the sequence.
In general, when executing a program containing comprehension
operators, we precompute the results of applying $\mcompr(f,i)$ for
each $f$ appearing as the first argument of $\mcompr$, where the number $i$
ranges from $0$ to $n_\mcompr-1$.  The number $n_\mcompr$ is a parameter called 
the \emph{comprehension limit}.  The pre-computation times out if it takes more 
than $i \times t_{\mathit{call}}$ time units to produce the outputs for 
$\mcompr(f,0), \ldots,\mcompr(f,i)$. 
When the top program is executed, a call to a $\mcompr(f,a)$ subprogram times out if no precomputed value exists for the input $i$ created by the subprogram $a$. Otherwise, the call returns the precomputed value for $\mcompr(f,i)$ to the top program.

\subsection{Choice of the Timeout Parameters}\label{sec:timeout}
The two parameters that determine how long a program may be run for are
the \emph{timer per call} $t_{\mathit{call} }$ and the \emph{comprehension
limit} $n_\mcompr$. A program times out if it exceeds the timeout or
if one of the $\mcompr$ expressions reaches the comprehension limit or if an 
integer 
with an absolute value larger than or equal to $2^{1025}$ is produced.  We may
run either a \emph{fast check}, a \emph{slow check} or a \emph{hybrid
check} on the set of candidate programs.  The fast check uses as
parameters $t_{\mathit{call}}=1000$, $n_\mcompr =20$, and the slow check uses
$t_{call}=100000$, $n_\mcompr=200$.

\paragraph{Hybrid Check}
The hybrid check tries to achieve the performance of the fast check
while retaining most of the additional solutions found by the slow check. 
The first phase of the hybrid check is the fast check. After this check, we
look at the programs that generated a prefix of an OEIS sequence but
could not complete the full task. At this point, if we were to perform
the slow check on all those prefix-generating programs, the hybrid
check would take a time equivalent to the slow check. To get a gain in
performance, we select only the ones that are the smallest for each
prefix to be tested for a longer time. Fast programs implicitly differentiate 
themselves from others by generating longer prefixes, therefore are also 
selected by the same criteria for further checking.

In most of our experiments, we use the hybrid check because it is
about 15 times faster than the slow check. However, since it does not
test all programs with the long timeout, it misses out on some
solutions found by the slow check.  In the longer NMT runs, we
eventually switched from the hybrid check to the more robust slow
check, to discover more solutions.

\section{OEIS Synthesis as an NMT Task}\label{sec:nmttask}

Neural networks have in the last decade
become competitive in language modeling and machine translation tasks,
leading to applications in many areas.  In particular, recurrent
neural networks (RNNs) with attention~\cite{ChorowskiBSCB15} and transformers~\cite{DBLP:conf/nips/VaswaniSPUJGKP17} have been
recently applied in mathematical and symbolic tasks such as rewriting~\cite{abs-1911-04873},
autoformalization~\cite{WangKU18} and synthesis of mathematical conjectures and proof
steps. Many of these tasks are naturally formulated as
sequence-to-sequence translation tasks.

\begin{figure}
    \centering
    \includegraphics[width=0.7\textwidth]{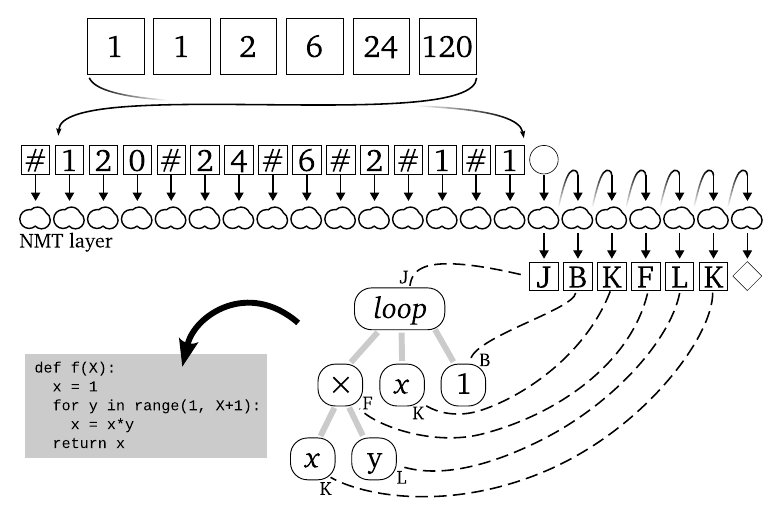}
    \caption{Representing sequences and solutions for NMT.}
    \label{fig:exnmt}
\end{figure}

\paragraph{NMT Representation}
The OEIS program synthesis can also be cast as such a task.  In this
work we therefore experiment with replacing the TNN architecture with a
reasonably fast encoder-decoder neural machine translation (NMT)
system.  In particular, we represent the input integer sequence as a
series of digits, separated by an additional token at the integer
boundaries. Since the initial integers in a sequence are typically
smaller and may be more informative for the NMT decoding phase, we
reverse the input sequences.  The output program is also represented
as a sequence of tokens in %
Polish notation (Fig.~\ref{fig:exnmt}).

\paragraph{Beam Search}
To make full use of the NMT capabilities, we also replace the original
MCTS search with a wide \emph{beam search} during the NMT
decoding. Beam search with width $N$ is an alternative to
\emph{greedy decoding}. Instead of a single greedily best output, NMT
in beam search keeps track of the $N$ conditionally most
probable outputs, updating their ranking after each decoding
step. When the NMT decoding for a particular input OEIS sequence $s$
is finished, the final $N$ best outputs can be used as NMT's $N$
alternative suggestions of programs that solve $s$. %

\paragraph{NMT Framework and Hyperparameters}
After several initial evaluations we have chosen for the experiments
Luong's NMT framework~\cite{luong17}. It works efficiently on our
hardware both in the training and wide-beam inference mode, and we
were able to find suitable hyperparameters for it~\cite{WangKU18}. In
more detail, we use for most experiments a 2-layer bidirectional LSTM
equipped with the ``scaled Luong'' attention and 512 units.  In our
NMT experiments (Section~\ref{sec:exp-nmt}) we start by using one NMT
model for training and inference, using many default NMT
hyperparameters. As the iterations progress, we gradually adjust the
parameters and add more models trained differently and on differently
selected data. We also experiment with larger models.

\paragraph{Combining NMT Models} In most NMT runs we %
train two to four different NMT
models in parallel %
each on its own
GPU. We then run the inference with two of them in parallel, thus using all 4 GPUs
on the server.
The rationale
behind training and inference with differently trained models is the
standard portfolio argument, used routinely, e.g., in automated
theorem proving~\cite{Tammet98,blistr,SchaferS15,JakubuvU18a,DBLP:conf/mkm/HoldenK21}.
A complementary portfolio of
specialists typically outperforms a single general
strategy. In feedback loops that
alternate between proof search and learning~\cite{US+08}, this also further
benefits the learning phase, since each learner can
additionally use the training data accumulated by others.
In Section~\ref{sec:exp-nmt} we see that this indeed considerably improves the performance.

\paragraph{Continuous training}
NMT models can be trained either only on the latest version of the solutions or 
in a continuous way. The latter method re-uses the model trained in
the previous iteration and trains it on the latest data for more
steps. This makes such model more stable, being eventually trained
for orders of magnitude more steps.  It also makes it different from the models trained from
scratch only on the latest data.  Even
when only a few solutions arrive in the latest iteration, the
network is training further on the whole latest corpus, thus becoming
smarter and hopefully more competent for the next 
inference phase. The models trained only on the latest corpus are on the other hand
less concerned by the old (slower/longer) solutions and more focused
on exploring the latest ones.

\section{Experiments with a Tree Neural Network}\label{sec:tnn}

In our previous work~\cite{abs-2202-11908}, a tree neural network (TNN) serves 
as a machine learning model. In the following, we test how 
varying a range of parameters influences the self-learning process.
To test the limit of the TNN, we run the TNN-guided learning loop for 500 
generations instead of the original 25. This last %
experiment of this section also provides a 
baseline for the NMT experiments.

\paragraph{Varying Parameters}
We investigate the effect of three different parameters: the TNN embedding 
size, the choice of the programming language and the choice of the objectives. 
Unless specified otherwise, the default value for those parameters are respectively 96 for 
the dimension, the previously described programming language (see 
Section~\ref{sec:program}) and the selection of the smallest and fastest 
programs. 
\begin{table}
          \caption{Evolution of the number of solutions of runs with various 
          parameters over the first 20 iterations}\label{tab:varying}
		\begin{center}
			\begin{tabular}{l|l|rrrrr}
						\toprule & Iteration & 0 & 5 & 10 & 15 & 20\\
				
	            \midrule
				&\textit{16} & 2005 & 14920 & 19674 & 21163 & 22203\\
				&\textit{32} & 1972 & 17608 & 23490 & 25750 & 27351\\
				Embedding size &\textit{64} & 2017 & 18156 & 23737 & 26490 & 
				28463\\  
				&\textit{96} & 1993 & 16051 & 20127 & 22890 & 24718\\
				&\textit{96} l.s. & 3771 & 20434 & 25378 & 28361 & 30344\\

				\midrule
				& \textit{minimal} & 1157 & 7096 & 8547 & 9126 & 9965\\
				Language (embed. 64) & \textit{default} & 2017 & 18156 
				& 23737 & 26490 & 28463\\  
				& \textit{extra} & 1763 & 18690 & 23757 & 26905 & 28794\\
				\midrule
				&\textit{both} & 3771 & 20434 & 25378 & 28361 & 30344\\
			Objective (local search) &\textit{small} & 3725 & 20501 & 26231 & 
			29124 & 31520\\
				&\textit{fast} & 3716 & 21021 & 26270 & 29326 & 31421\\
				\bottomrule
			\end{tabular}
		\end{center}
            \end{table}
In Table~\ref{tab:varying}, we present the results of running experiments with 
different parameters for 20 iterations.
In a first experiment (first block in the table), we observe that the number of 
solutions increases with the dimension until dimension 64. Surprisingly, 
dimension 96 gave worse results, this is mainly due to the fact that larger 
networks are more expensive to compute and therefore produce fewer programs. As 
a countermeasure, we introduce a local search (l.s.) that tests all programs 
that are one action away from being constructed in the search tree. This 
balances the generation time and checking time better, leading to an increase 
in the number of solutions.
In a second experiment (second block in the table), we measure how changing the 
programming language affects the performance. All the experiments presented in 
this block use dimension 64. The \textit{minimal} row runs the self-learning 
loop with an even more minimalistic programming language consisting of four 
operators: $0,\mathit{successor},\mathit{predecessor},\mloop$. This makes the
 learning much more challenging. One of the 
reasons is that creating a large number such as one million in this language 
requires at least one million steps. Therefore, it is 
impossible to produce large numbers within the checker's time limit.
Such considerations justify the inclusion in the default language of operators efficiently computed 
by current hardware such as $+$ and $\times$.
The \textit{extra} row shows what happens when we include the extra constants 
3,4,...,10 as primitive operators. This inclusion makes the computation of 
large numbers more efficient and increases the performance of our system 
slightly. However, adding more and more primitive operators introduces human 
bias that we 
would rather avoid. 
In a third experiment (third block in the table), the results of learning with 
different program objectives are presented. All these experiments were 
performed with dimension 96 and local search. The 
\textit{small} (resp. \textit{fast}) row shows the effect of only collecting 
and learning from the smallest (resp. the fastest) programs for a given OEIS 
sequence. The results imply that focusing on one 
objective at a time simplifies the work of the machine learner. Yet, we expect more synergy between the two objectives to occur during longer runs.

\pgfplotscreateplotcyclelist{rw}
{solid, mark repeat = 10, mark phase = 10000, mark = *, black\\
	solid, mark repeat = 10, mark phase = 200, mark = square*, black\\
	solid, mark repeat = 10, mark phase = 200, mark = square*, black\\
	solid, mark repeat = 10, mark phase = 200, mark = square*, black\\}

\paragraph{Long Run}

The result of a long-lasting experiment, running for 500 iterations with the 
default parameters and local search, is shown in Fig.~\ref{fig:allcummul} (tnn). %
To fit also the NMT runs, we display only the first 190 iterations, however the average increments (Fig.~\ref{fig:allincr}, tnn) between iteration 200 and 300 drops below 20. At 
the end of the run, the TNN seems to have reached its limit and about five new 
solutions are found at each iteration. %
As we will see in Section~\ref{sec:exp-nmt}, due to its larger embedding size 
and its more involved architecture and the introduction of continuous training, the main NMT experiment does not plateau and
reaches a much higher number in an equivalent amount of time.

\section{Experiments with NMT}~\label{sec:exp-nmt}
\subsection{Basic run (\texorpdfstring{$\mathit{nmt}_0$}{nmt0})}
In the basic NMT
run\footnoteA{\url{https://github.com/Anon52MI4/oeis-alien/tree/master/run0}}
($\mathit{nmt}_0$), we are
running the loop in a way that is most similar to the TNN run. In
particular, the checking phase is interleaved with training a single
NMT $\mathit{model}_0$, which is then used for the search phase, implemented as NMT
inference using a wide beam search. As in the TNN experiment, we start with the initial random
iteration which yields 3771 solutions. Then we run 100 iterations of
the NMT-based learn-generate-check loop.

\paragraph{Training} We use a batch size of 512 and SGD with a learning 
rate of 1.0. In each iteration, we train on the latest set of solutions, 
for 12000 steps. From iteration 45, we increased the number of training steps 
to 14000 to adjust for the growing number of training examples.\footnoteA{This 
corresponds to
438 epochs in iteration 3 where there are about 14k
examples, 92 epochs in iteration 44 (67k examples), 107 epochs in iteration 45 
(after switching to 14000 steps),
and 76 epochs in iteration 101 (81k examples).
On one GPU (GTX1080 Ti, 12G RAM) the training 
takes on average 100m with 12000 steps and 130m with 14000 steps.}
The bleu scores on small test and development sets are typically between 25 and 30.
There is only one memory crash (iteration 97). %
\paragraph{Inference} We use two GPUs in parallel (splitting OEIS into
two parts), each with a batch size of 32 and beam width 240.  These
values are determined experimentally, to load the GPUs efficiently
without memory crashes. The parallelized inference time grows from 2
to 8 hours, increasing as the iterations invent longer examples, and
the trained network and the beam search become less confident about
when to stop decoding. Each inference phase yields %
$240 \cdot 351663 = 84.4M$ program candidates that are then %
handed over to the checking phase. %
\paragraph{Checking} We use the hybrid checking mode
(Section~\ref{sec:timeout}) parallelized over 18 CPUs. The
checking time grows from about 2 minutes to about 8-10 minutes. This is
negligible compared to the NMT training and inference
times. The 100 iterations of this loop took about a month of real time, 
reaching 46707 solutions (Fig.~\ref{fig:allcummul}, nmt0). %
However, the increments drop below 200 (respectively 100) after iteration 37 
(respectively 94).

\subsection{Long extended run (\texorpdfstring{$\mathit{nmt}_1$}{nmt1})}
Since the time taken by one $\mathit{nmt}_0$ iteration reaches about half a day towards the end of the run, we explore
more efficient approaches and further extensions and modifications. %
This leads to the longest %
run\footnoteA{\url{https://github.com/Anon52MI4/oeis-alien/tree/master/run1}} $\mathit{nmt}_1$, which has at the time of submitting this paper reached
190 iterations and over 78000 solutions (Fig.~\ref{fig:allcummul}, nmt1).\footnoteA{The 190 $\mathit{nmt}_1$ iterations took 3 months on a 4-GPU server.} Its remarkable property is that,
unlike in $\mathit{nmt}_0$, the number of new solutions produced in each
iteration rarely drops below 200, even after many iterations (Fig.~\ref{fig:allincr}, nmt1). This
challenges the standard wisdom of ``plateauing curves'' appearing in the TNN 
and $\mathit{nmt}_0$ runs, and in many similar search-learn feedback loops~\cite{KaliszykUMO18,PiotrowskiU18,JakubuvU19}.

\paragraph{Combining models}
The $\mathit{nmt}_1$ run bootstraps from $\mathit{nmt}_0$, inheriting
its first 20 iterations, thus starting with %
34420 solutions. %
Then we start combining multiple NMT models (Section~\ref{sec:nmttask}).
Since iteration 21, we %
add training of $\mathit{model}_1$ to $\mathit{model}_0$. It is trained only on a
randomly chosen half of the training set, however for twice as many
steps/epochs. %
This yields a %
differently trained (and more focused) specialist in each
iteration. The number of solution candidates produced by the inference
phase thus doubles, to 168.8M. After de-duplication, this yields 32.5M
unique candidates %
in iteration 21 %
compared to 13.4M in iteration 20. The checking (initially
also using the hybrid mode) is still fast, taking 5 minutes on
18 CPUs. The difference to $\mathit{nmt}_0$ %
is remarkable: 687 new solutions in $\mathit{nmt}_1$ vs 272 in $\mathit{nmt}_0$ 
in iteration 21. This effect continues over the next iterations, see 
Fig.~\ref{fig:allincr}.
\paragraph{Further modifications}
Since iteration 70, we start training $\mathit{model}_1$ in a continuous way 
(Section~\ref{sec:nmttask}).
By iteration 190 $\mathit{model}_1$  is thus trained for over
1.4M steps on the evolving data, making it quite 
different from other models.
As we invent longer programs, we also allow training on
longer sequences, raising the default NMT values of 50 input
and 50 output tokens gradually to 80 and 140, respectively.
Since iteration 156 we also add training of $\mathit{model}_2$, %
which takes %
specialization even further.
It is trained for four times as many steps as $\mathit{model}_0$ on a random quarter
of the data. The bleu scores of $\mathit{model}_0$  are at
this point low, due to the larger size of the data, while
$\mathit{model}_2$ still achieves scores above 25. %
We then use $\mathit{model}_0$ only as a backup %
when $\mathit{model}_2$ diverges. %
Since iteration 159 we switch from the hybrid check to the slow (full)
check (Section~\ref{sec:timeout}), raising the checking time from 45m
(iteration 158) to 6h, and the number of solutions from 178 to 860
(Fig.~\ref{fig:allincr}, nmt1). This jump is likely due to many
programs using comprehension being newly allowed. It includes sudden
solutions for hundreds of problems that combine congruence operations and
primes.\footnoteA{\url{https://bit.ly/3QPkquE}}

\paragraph{Bigger Network}
Since iteration 170, we add continuous training of a bigger $\mathit{model}_3$ 
which uses
1024 units instead of 512. To keep all models and phases in 
sync, we train $\mathit{model}_3$ for fewer steps (8000), decreasing also its batch size and inference width to 288 and 120, respectively.
Table~\ref{tab:mods} analyzes the benefits of using $\mathit{model}_3$, $\mathit{model}_2$ and $\mathit{model}_0$ in addition to $\mathit{model}_1$over 12 iterations (175-186).
\begin{table} %
  \caption{\label{tab:mods} Influence of using models M3, M2 or M0 for inference in addition to M1. UC are unique candidates (millions), NS new solutions added, TS total solutions including all found so far, and OS own solutions, i.e., the sequences covered by the current iteration.}
\begin{center}
  \begin{tabular}{lllllllllllll}
    \toprule
    \emph{Iteration} & \emph{175} & \emph{176} & \emph{177} & \emph{178} & 
    \emph{179} & \emph{180} \\
    Model & M2& M3 & M0 & M2 & M2 & M2 \\
     UC         & 101.7 & 79.7 & 64.7 & 99.7 & 96.0 & 106.2 \\
     OS         & 74746 & 74916 & 74130 & 75196 & 75386 & 75313 \\
     TS         & 75471 & 75666 & 75795 & 75985 & 76160 & 76404 \\
     NS         & 260 & 195 & 129 & 190 & 175 & 244 \\
    \midrule
    \emph{Iteration} & \emph{181} & \emph{182} & \emph{183} & \emph{184} & 
    \emph{185} & \emph{186} \\
    Model & M2 & M2 & M2 & M2 & M3 & M3 \\
    UC & 105.8 & 106.1 & 106.7 & 101.1 & 84.5 & 85.7 \\
     OS         & 75686 & 75986 & 76271 & 76397 & 76793 & 77007 \\
     TS         & 76656 & 76861 & 77055  & 77244 & 77411 & 77577 \\
     NS         & 252 & 205 & 194 & 189 & 167 & 166 \\
    \bottomrule
  \end{tabular}
\end{center}
\end{table}
The number of unique candidates is much lower for $\mathit{model}_3$ (due to the beam width) than for $\mathit{model}_2$, however still higher than for $\mathit{model}_0$, which is at this point likely undertrained. In general, $\mathit{model}_3$ performs better than $\mathit{model}_0$, but is inferior to $\mathit{model}_2$. A faster model, producing twice as many
plausible candidates in the same amount of time, is here better than a slower
model which is a bit more precise.\footnoteA{Also, $\mathit{model}_3$ is trained
continuously and may resemble more $\mathit{model}_1$, providing fewer different candidates.
And $\mathit{model}_2$ is trained only on a random quarter of the
latest data, making it even more orthogonal to the continuous models
that have seen much more data.}  

\subsection{Runs with Local and Global Macros}
\begin{figure}[t]
	\centering
	\begin{tikzpicture}[scale=0.9]
	\begin{axis}[
	legend style={anchor=north west, at={(0.1,0.9)}},
	width=\textwidth,
	height=0.65*\textwidth,xmin=20, xmax=190,
	ymin=0, ymax=60,
	cycle list name=rw,
	xtick = {0,25,50,75,100,125,150,175},
	scaled y ticks = false,
	ylabel style={yshift=0pt},
	xlabel = {Iteration},
	ylabel = {Average size} 
	]
	\addplot table [x=gen, y=small] {1sizesnan};
	\addplot table [x=gen, y=fast] {1sizesnan};
	\legend{small, fast};
	\end{axis}
	\end{tikzpicture}
	\caption{Average size of the smallest (respectively fastest) programs 
	         discovered within $x$ iterations}
	\label{fig:avrgsize}
\end{figure}
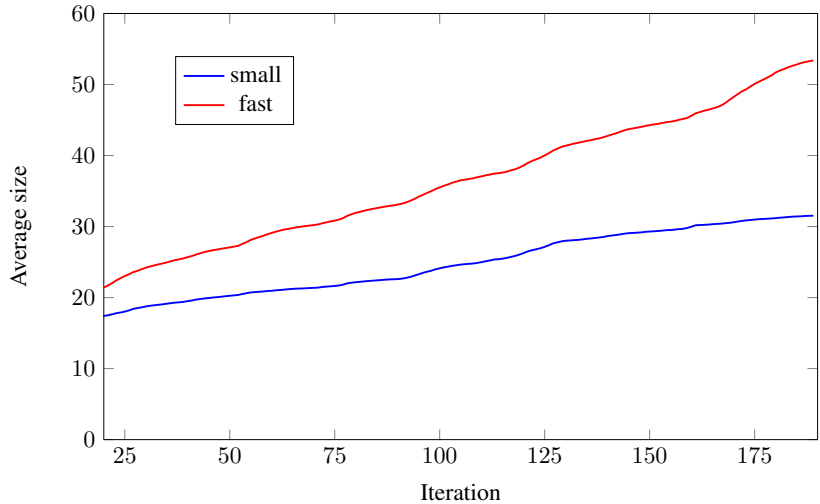

As the average program size grows in $\mathit{nmt}_1$
(Fig.~\ref{fig:avrgsize}), the NMT decoding times increase, taking
almost 12h in the latest iterations. Verbatim repetition of blocks of
code also feels suboptimal to human programmers. This motivates later 
additional runs $\mathit{nmt}_2$ and $\mathit{nmt}_3$, where we experiment with 
using local and global macros. Apart from allowing these additional macro 
mechanisms (Section~\ref{sec:program}), the two runs resemble run 
$\mathit{nmt}_1$. In each of them we train the basic $\mathit{model}_0$ 
together with the continuous $\mathit{model}_1$, and infer with both of them. 
Since the sequences are shorter (making $\mathit{model}_0$ easier to train), we 
have not so far experimented with $\mathit{model}_2$ here. While the two runs 
seem superior to $\mathit{nmt}_1$ in Fig.~\ref{fig:allincr}, this is mainly due 
to the use of both models since iteration 2 (instead of iteration 21 in 
$\mathit{nmt}_1$), and also an earlier switch to the slow check (iteration 75 
in $\mathit{nmt}_2$ and iteration 67 in $\mathit{nmt}_3$). The two runs also do 
not significantly complement $\mathit{nmt}_1$: each of them adds less than 2800 
solutions to $\mathit{nmt}_1$. This may mean that the alien system 
$\mathit{nmt}_1$ has so far less trouble than humans with expanding everything 
and the use of definitions is not as critical as in humans. None of the two 
runs has however reached iteration 100 yet, making the comparison with 
$\mathit{nmt}_1$ only preliminary. Especially the statistics of the use of 
global macros\footnoteA{\url{https://bit.ly/3ZNe7fm}} in $\mathit{nmt}_3$ are 
interesting, and can be used for analyzing the evolution of its coding trends.

\pgfplotscreateplotcyclelist{rw}{
solid, thick, blue\\
solid, thick, red\\
solid, thick, yellow!50!orange\\
solid, thick, green\\
solid, thick, orange\\
}

\begin{figure}[t]
	\centering
	\begin{tikzpicture}[scale=0.9]
	\begin{axis}[
	legend style={anchor=north west, at={(0.1,0.9)}},
	width=\textwidth,
	height=0.65*\textwidth,xmin=0, xmax=190,
	ymin=0, ymax=80000,
	cycle list name=rw,
	xtick = {0,25,50,75,100,125,150,175},
	scaled y ticks = false,
	ylabel style={yshift=10pt},
	xlabel = {Iteration},
	ylabel = {Sequences} 
	]
    \addplot table [x=gen, y=tnn, col sep=tab] {1allcumulnan};
    \addplot table [x=gen, y=nmt0, col sep=tab] {1allcumulnan};
    \addplot table [x=gen, y=nmt1, col sep=tab] {1allcumulnan};
    \addplot table [x=gen, y=nmt2, col sep=tab] {1allcumulnan};
    \addplot table [x=gen, y=nmt3, col sep=tab] {1allcumulnan};
	\legend{tnn, nmt0, nmt1, nmt2, nmt3};
	\end{axis}
	\end{tikzpicture}
	\caption{Number $y$ of solutions after $x$ iterations}
	\label{fig:allcummul}
\end{figure}
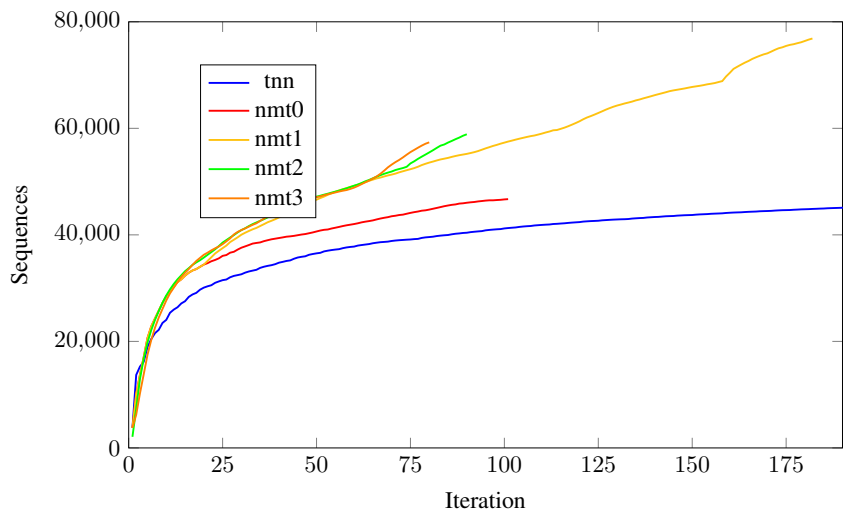

\begin{figure}[t]
	\centering
	\begin{tikzpicture}[scale=0.9]
	\begin{axis}[
	legend style={anchor=north west, at={(0.1,0.9)}},
	width=\textwidth,
	height=0.65*\textwidth,xmin=20, xmax=190,
	ymin=0, ymax=1000,
	cycle list name=rw,
	xtick = {0,25,50,75,100,125,150,175},
	scaled y ticks = false,
	ylabel style={yshift=0pt},
	xlabel = {Iteration},
	ylabel = {Sequences} 
	]
	\addplot table [x=gen, y=tnn, col sep=tab] {1allincrnan};
	\addplot table [x=gen, y=nmt0, col sep=tab] {1allincrnan};
	\addplot table [x=gen, y=nmt1, col sep=tab] {1allincrnan};
	\addplot table [x=gen, y=nmt2, col sep=tab] {1allincrnan};
	\addplot table [x=gen, y=nmt3, col sep=tab] {1allincrnan};
	\legend{tnn, nmt0, nmt1, nmt2, nmt3};
	\end{axis}
	\end{tikzpicture}
 \caption{Number of new solutions $y$ found at each iteration $x$.}
\label{fig:allincr}
\end{figure}
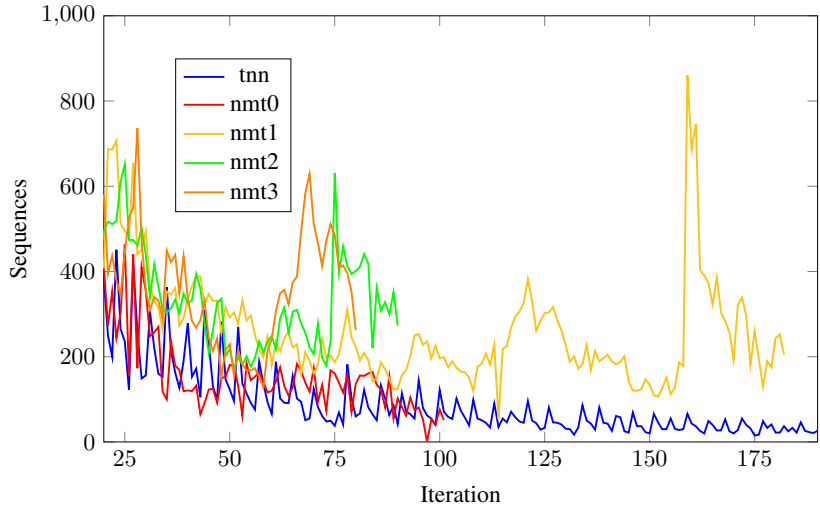

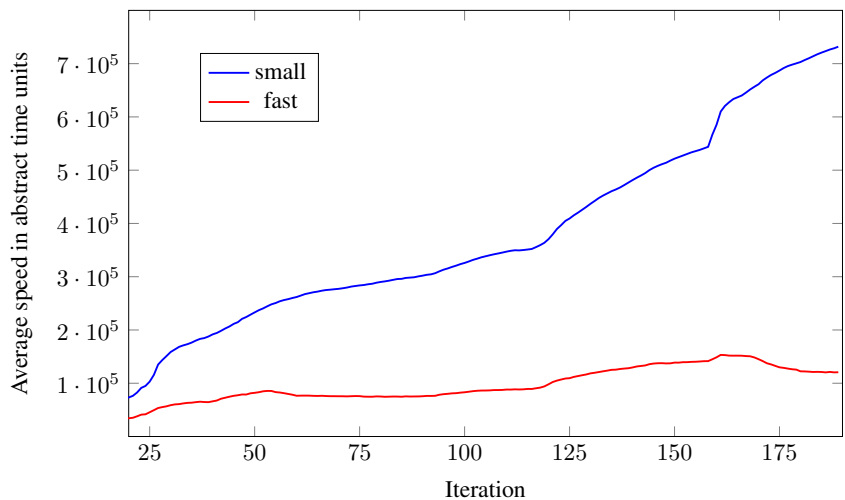
\begin{figure}[t]
	\centering
	\begin{tikzpicture}[scale=0.9]
	\begin{axis}[
	legend style={anchor=north west, at={(0.1,0.9)}},
	width=\textwidth,
	height=0.65*\textwidth,xmin=20, xmax=190,
	ymin=0, ymax=800000,
	cycle list name=rw,
	xtick = {0,25,50,75,100,125,150,175},	
	ytick = {100000,200000,300000,400000,500000,600000,700000},
	scaled y ticks = false,
    ylabel style={yshift=10pt},
	xlabel = {Iteration},
	ylabel = {Average speed in abstract time units} 
	]
	\addplot table [x=gen, y=small] {1timesnan};
	\addplot table [x=gen, y=fast] {1timesnan};
	\legend{small, fast};
	\end{axis}
	\end{tikzpicture}
	\caption{Average speed of the fastest (respectively smallest) 
	programs discovered within $x$ iterations}
	\label{fig:avrgtime} 
\end{figure}

\begin{figure}[ht]
	\centering
	\begin{tikzpicture}[scale=0.9]
	\begin{axis}[
	legend style={anchor=north west, at={(0.1,0.9)}},
	width=\textwidth,
	height=0.65*\textwidth,xmin=0, xmax=100,
	ymin=21, ymax=26,
	cycle list name=rw,
	xtick = {0,25,50,75,100,125,150,175},
	scaled y ticks = false,
	ylabel style={yshift=0pt},
	xlabel = {Additional iterations},
	ylabel = {Average size} 
	]
	\addplot table [x=gen, y=size] {1comprsizenan};
	\end{axis}
	\end{tikzpicture}
	\caption{Evolution of the average size $y$ of smallest programs for 100
	additional iterations starting from the iteration they were first 
	discovered}
	 \label{fig:sizecompr}
\end{figure}
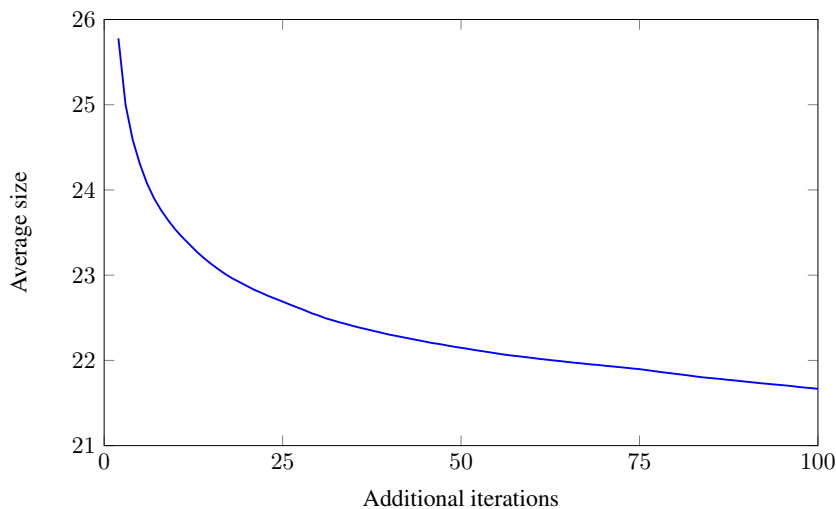

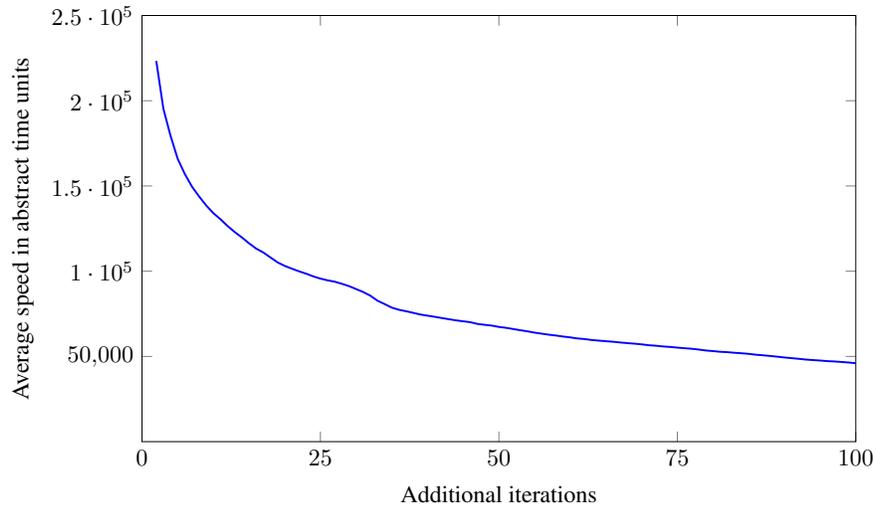
\begin{figure}[ht]
	\centering
	\begin{tikzpicture}[scale=0.9]
	\begin{axis}[
	legend style={anchor=north west, at={(0.1,0.9)}},
	width=\textwidth,
	height=0.65*\textwidth,xmin=0, xmax=100,
	ymin=0, ymax=250000,
	cycle list name=rw,
	xtick = {0,25,50,75,100,125,150,175},	
	ytick = {50000,100000,150000,200000,250000,300000},
	scaled y ticks = false,
	ylabel style={yshift=15pt},
	xlabel = {Additional iterations},
	ylabel = {Average speed in abstract time units}
	]
	\addplot table [x=gen, y=time] {1comprtimenan};
	\end{axis}
	\end{tikzpicture}
	\caption{Evolution of the average size $y$ of fastest programs for 100
		additional iterations starting from the iteration they were first 
		discovered}
	\label{fig:timecompr}
\end{figure}

\section{Analysis of the Results}\label{sec:analysis}
We provide %
a statistical analysis of the 78118 solutions found during the 
$\mathit{nmt}_1$ run and discuss the details of some %
techniques
developed by our system. More information on the $\mathit{nmt}$ runs is 
available in our anonymous 
repository.\footnoteA{\url{https://bit.ly/3iVIfnX}} For some sequences,
its subdirectory\footnoteA{\url{https://bit.ly/3XHZsjK}} contains our analysis 
of the 
evolution\footnoteA{\url{https://bit.ly/3iJ4oGd}}\footnoteA{\url{https://bit.ly/3HfwemI}}
and 
proliferation\footnoteA{\url{https://bit.ly/3ZNExO4}}\footnoteA{\url{https://bit.ly/3IVzrJz}}
of important programs such as primes and sigma, as well as
proofs that some of
the alien programs match the human OEIS intention.\footnoteA{\url{https://bit.ly/3QPB25o}}\footnoteA{\url{https://bit.ly/3WlVHPM}}\footnoteA{\url{https://bit.ly/3XJi96j}}
See also the appendix for more details.
Noteworthy sequences from this 
repository include convolution of primes with 
themselves,\footnoteA{\url{https://bit.ly/3XJi96j}} showing the proficiency of 
our system with primes. Motzkin 
numbers\footnoteA{\url{https://bit.ly/3QPB25o}}~\cite{DBLP:journals/jintseq/WangZ15}
are an example where our synthesized programs rely on a \emph{pairing 
function}. %
This programming technique, re-invented by our system, packs two variables 
into one, allowing further programs (see Appendix~\ref{sec:triangle} for details).
A solution for the unique monotonic sequence of 
nonnegative integers satisfying $a(a(n)) = 3n$ is needed in solving 
a problem in $27^{th}$ British Mathematical Olympiad.\footnoteA{\url{https://bit.ly/3wjmqSg}}

\paragraph{Evolution of the Programs}
Fig.~\ref{fig:avrgsize} shows the evolution of the average size
and Fig.~\ref{fig:avrgtime} the speed  of the
solutions. We see that as new
solutions are found, they become longer and typically also take more
time. However, there are interesting exceptions to the latter rule
(Fig.~\ref{fig:avrgtime}), as more efficient code is invented and
propagated by the alien system. We thus also measure the gradual size
reduction (Fig.~\ref{fig:sizecompr}) and time reduction
(Fig.~\ref{fig:timecompr}) for the short and fast solutions
(respectively). This is for each sequence computed for 100 iterations
after its first solution was found. We see that the iterations induce a
remarkable %
speedup of the invented fast solutions (Fig.~\ref{fig:timecompr}).

\paragraph{Generalization of the Solutions to Larger Indices}
OEIS provides additional terms for some of the OEIS entries in b-files.
Among the 78118 solutions, 40,577 of them have a b-file that contains 100 additional terms for their OEIS entry.
We evaluate both the small and the fast programs with the slow check parameters on the 100 additional terms.
Here, 14,701 small and 11,056 fast programs time out.
Among the programs that do not time out, the percentage of generalizing programs (producing matching additional terms) 
is 90.57\% for the slow programs and 77.51\% for the fast programs.
A common %
error is %
reliance on an approximation for a real number, 
such as $\pi$. 

\section{Related work}\label{sec:related}

The closest related recent work is~\cite{DBLP:journals/corr/abs-2201-04600},
done in parallel to our project.
The goal there is similar to ours but their 
general approach is focused on training a single model
using supervised learning techniques on synthetic data.
In contrast, our approach is based on reinforcement learning: we start from 
scratch and keep training new models 
as we discover new OEIS solutions.
The programs generated in~\cite{DBLP:journals/corr/abs-2201-04600} are 
recurrence relations defined by analytical formulas.
Our language seems to be more expressive. For example, our language can use the 
functions it defines by recurrence using $\mloop$ and $\mloopt$ as subprograms 
and construct nested loops.
The performance of the model in ~\cite{DBLP:journals/corr/abs-2201-04600} is 
investigated on real number sequences, whereas our work focuses only on integer 
sequences. Overall, it is hard to directly compare the performance of the two 
systems. Our result is the number of OEIS sequences found by targeting the 
whole encyclopedia, whereas~\cite{DBLP:journals/corr/abs-2201-04600} reports 
the 
test accuracy only on 10,000 easy OEIS 
sequences.
A related recent small-scale experiment is reported in~\cite{bwcw22},
where a model pre-trained on many code repositories is used (i.e., the
system is not developed from scratch as we investigate here), using a
language and feedback-loop setting similar to the one introduced by us
in \cite{abs-2202-11908}. The reported result there is 11.5\% of 10000
easy OEIS sequences. This is also incomparable to our system, where
the main point is to measure the capability of the system to develop
new ideas on its own (i.e., without the potential contamination with
ideas seen in the pre-training), and the system runs on the whole
OEIS, reporting orders of magnitude higher numbers of solutions.

The deep reinforcement learning system %
\textsf{DreamCoder\xspace}~\cite{DBLP:conf/pldi/EllisWNSMHCST21} %
has demonstrated self-improvement from scratch in several programming
tasks. Its main contribution is the use of definitions that compress
existing solutions and facilitate building new solutions on top of the
existing ones.  Our experiments with local and global macros are
however so far inconclusive. While humans certainly benefit from
introducing new names, concepts and shortcuts, it is so far unclear if it
results in a clear improvement in our current setting.
Another related system is developed within the LODA~\cite{loda} project, where the methods, experiments and resources are however yet to be fully described. Within the theorem proving community, the development of methods for 
term synthesis has been explored in inductive theorem 
proving~\cite{hipspec13} and in counterexample 
generators~\cite{DBLP:conf/itp/BlanchetteN10,DBLP:conf/cpp/Bulwahn12}.

While the previous version of our system used MCTS inspired by
AlphaGoZero~\cite{silver2017mastering}, the current NMT approach uses
just straightforward beam search. The general search-verify-learn
positive feedback loop between ML-guided symbolic search and
statistical learning from the verified results has been used for a long
time in large-theory automated theorem proving, going back at least to
the MaLARea system~\cite{Urban07,US+08}. Its recent instances include
systems such as rlCoP~\cite{KaliszykUMO18} and
ENIGMA~\cite{JakubuvU19}. Such systems can be also seen as synthesis
frameworks, where the mathematical object synthesized by the guided
search is however the full proof of a theorem, rather than just a
particular interesting witness or a conjecture. Synthesis of full
proofs of mathematical theorems is an
all-encompassing task that can be extremely hard in cases like
Fermat's Last Theorem. Our current setting may thus also be seen as an attempt
to decompose this all-encompassing task into smaller interesting
subtasks that can be analyzed separately.

\section{Conclusion}
As of January 25, 2023, all of the runs have together invented from scratch solutions for 84587 OEIS sequences.
This is more than three times the number (27987) invented in our first
experiments~\cite{abs-2202-11908}.  This is due to several
improvements. We have started to collect both the small and fast
solutions and learn jointly from them. We have seen that this
gradually leads to a large speedup of the fast programs, as the
population of programs evolves. This likely allows invention of
further solutions that are within our time limit, thanks to the discovery of
the faster and faster components. The improvements (learning)
on the symbolic side thus likely complement and co-evolve with the
statistical learning used for guiding the synthesis.
We have also
started to use relatively fast neural translation models, their
specialization on random subsets, their combinations and continuous
training, and a wide beam search instead of MCTS. The experiments
suggest that these techniques are useful, leading to a high rate of
program invention even after the 190 iterations in the longest NMT
run.  This is encouraging since many of the solved OEIS problems seem
nontrivial. The trained system could thus be used as a search tool
assisting mathematicians.  There is also a wealth of related
mathematical tasks that can be cast in a similar synthesis setting,
and combined with other tools, such as automated theorem provers.

A crucial element of our setting is the fact that NMT is used to
produce an interpretable symbolic representation. It would be
practically unusable if we relied purely on neural approximation of
arbitrary functions and the task for NMT was to directly produce the
next 100 numbers in each OEIS sequence.  The interpretable symbolic
representation is critical for the generalization capability of the
overall system and its capability to learn from itself.
The overall alien system can also be seen as an example of a very weakly
supervised evolutionary architecture where simple high-level
principles (Occam's razor and efficiency) govern the development and
training of the lower-level statistical component (NMT). Rather than
one-time training on everything that has been already invented by
humans so far, as done by today's large language models, the system
here starts from zero knowledge and progresses towards increasingly
nontrivial knowledge and skills.  Such feedback loops thus seem to be
a good playground for exploring how increasingly intelligent systems
emerge. Note that thanks to the Turing completeness of the language,
this particular playground is (unlike games like Chess and Go) not
limited in its expressivity, and can in principle lead to the
development of arbitrary complex algorithms.

\section{Acknowledgments}
We thank Chad Brown, David Cerna, Hugo Cisneros, Tom
Hales, Barbora Hudcova, Jan Hula, Mikolas Janota, Tomas Mikolov, Jelle
Piepenbrock, and Martin Suda for discussions, comments and suggestions.
This work was partially supported by the CTU Global Postdoc funding
scheme (TG), Czech Science Foundation project 20-06390Y (TG),
 ERC-CZ project POSTMAN no. LL1902 (TG), Amazon
Research Awards (TG, JU), EU ICT-48 2020 project TAILOR no. 952215 (JU),
and the European Regional Development Fund under the Czech project
AI\&Reasoning with identifier CZ.02.1.01/0.0/0.0/15\_003/0000466 (MO, JU).

\bibliographystyle{plain}
\bibliography{biblio,ate11}

\begin{thebibliography}{10}

\bibitem{andrychowicz2017hindsight}
Marcin Andrychowicz, Filip Wolski, Alex Ray, Jonas Schneider, Rachel Fong,
  Peter Welinder, Bob McGrew, Josh Tobin, OpenAI Pieter~Abbeel, and Wojciech
  Zaremba.
\newblock Hindsight experience replay.
\newblock {\em Advances in neural information processing systems}, 30, 2017.

\bibitem{DBLP:conf/itp/BlanchetteN10}
Jasmin~Christian Blanchette and Tobias Nipkow.
\newblock {Nitpick}: {A} counterexample generator for higher-order logic based
  on a relational model finder.
\newblock In {\em Interactive Theorem Proving, First International Conference,
  {ITP} 2010, Edinburgh, UK, July 11-14, 2010. Proceedings}, pages 131--146,
  2010.

\bibitem{DBLP:conf/cpp/Bulwahn12}
Lukas Bulwahn.
\newblock The new {Quickcheck} for {Isabelle} - random, exhaustive and symbolic
  testing under one roof.
\newblock In {\em Certified Programs and Proofs - Second International
  Conference, {CPP} 2012, Kyoto, Japan, December 13-15, 2012. Proceedings},
  pages 92--108, 2012.

\bibitem{bwcw22}
Natasha Butt, Auke Wiggers, Taco Cohen, and Max Welling.
\newblock Program synthesis for integer sequence generation.
\newblock \url{https://mathai2022.github.io/papers/24.pdf}, 2022.

\bibitem{ChorowskiBSCB15}
Jan Chorowski, Dzmitry Bahdanau, Dmitriy Serdyuk, Kyunghyun Cho, and Yoshua
  Bengio.
\newblock Attention-based models for speech recognition.
\newblock In {\em {NIPS}}, pages 577--585, 2015.

\bibitem{hipspec13}
Koen Claessen, Moa Johansson, Dan Ros{\'e}n, and Nicholas Smallbone.
\newblock Automating inductive proofs using theory exploration.
\newblock In Maria~Paola Bonacina, editor, {\em Conference on Automated
  Deduction (CADE)}, volume 7898 of {\em LNCS}, pages 392--406. Springer, 2013.

\bibitem{conway1970game}
John Conway et~al.
\newblock The game of life.
\newblock {\em Scientific American}, 223(4):4, 1970.

\bibitem{DBLP:journals/corr/abs-2201-04600}
St{\'{e}}phane d'Ascoli, Pierre{-}Alexandre Kamienny, Guillaume Lample, and
  Fran{\c{c}}ois Charton.
\newblock Deep symbolic regression for recurrent sequences.
\newblock {\em CoRR}, abs/2201.04600, 2022.

\bibitem{DBLP:conf/pldi/EllisWNSMHCST21}
Kevin Ellis, Catherine Wong, Maxwell~I. Nye, Mathias Sabl{\'{e}}{-}Meyer, Lucas
  Morales, Luke~B. Hewitt, Luc Cary, Armando Solar{-}Lezama, and Joshua~B.
  Tenenbaum.
\newblock {D}ream{C}oder: bootstrapping inductive program synthesis with
  wake-sleep library learning.
\newblock In Stephen~N. Freund and Eran Yahav, editors, {\em {PLDI} '21: 42nd
  {ACM} {SIGPLAN} International Conference on Programming Language Design and
  Implementation, Virtual Event, Canada, June 20-25, 2021}, pages 835--850.
  {ACM}, 2021.

\bibitem{abs-2202-11908}
Thibault Gauthier and Josef Urban.
\newblock Learning program synthesis for integer sequences from scratch.
\newblock {\em CoRR}, abs/2202.11908, 2022.

\bibitem{DBLP:conf/gcai/2015}
Georg Gottlob, Geoff Sutcliffe, and Andrei Voronkov, editors.
\newblock {\em Global Conference on Artificial Intelligence, {GCAI} 2015,
  Tbilisi, Georgia, October 16-19, 2015}, volume~36 of {\em EPiC Series in
  Computing}. EasyChair, 2015.

\bibitem{hamming1998mathematics}
Richard~Wesley Hamming.
\newblock Mathematics on a distant planet.
\newblock {\em The American Mathematical Monthly}, 105(7):640--650, 1998.

\bibitem{DBLP:conf/mkm/HoldenK21}
Edvard~K. Holden and Konstantin Korovin.
\newblock Heterogeneous heuristic optimisation and scheduling for first-order
  theorem proving.
\newblock In {\em {CICM}}, volume 12833 of {\em Lecture Notes in Computer
  Science}, pages 107--123. Springer, 2021.

\bibitem{JakubuvU18a}
Jan Jakub\r{u}v and Josef Urban.
\newblock Hierarchical invention of theorem proving strategies.
\newblock {\em {AI} Commun.}, 31(3):237--250, 2018.

\bibitem{JakubuvU19}
Jan Jakub\r{u}v and Josef Urban.
\newblock Hammering {Mizar} by learning clause guidance.
\newblock In John Harrison, John O'Leary, and Andrew Tolmach, editors, {\em
  10th International Conference on Interactive Theorem Proving, {ITP} 2019,
  September 9-12, 2019, Portland, OR, {USA}}, volume 141 of {\em LIPIcs}, pages
  34:1--34:8. Schloss Dagstuhl - Leibniz-Zentrum f{\"{u}}r Informatik, 2019.

\bibitem{KaliszykUMO18}
Cezary Kaliszyk, Josef Urban, Henryk Michalewski, and Miroslav Ols{\'{a}}k.
\newblock Reinforcement learning of theorem proving.
\newblock In {\em Advances in Neural Information Processing Systems 31: Annual
  Conference on Neural Information Processing Systems 2018, NeurIPS 2018, 3-8
  December 2018, Montr{\'{e}}al, Canada.}, pages 8836--8847, 2018.

\bibitem{loda}
Christian Krause.
\newblock {LODA}: An assembly language, a computational model, and a
  distributed tool for mining programs.
\newblock \url{https://github.com/loda-lang}, 2023.

\bibitem{langton1997artificial}
Christopher~G Langton.
\newblock Artificial life: An overview.
\newblock 1997.

\bibitem{luong17}
Minh{-}Thang Luong, Eugene Brevdo, and Rui Zhao.
\newblock Neural machine translation (seq2seq) tutorial.
\newblock {\em \url{https://github.com/tensorflow/nmt}}, 2017.

\bibitem{10.1093/comjnl/6.3.232}
Donald Michie.
\newblock {Experiments on the Mechanization of Game-Learning Part I.
  Characterization of the Model and its parameters}.
\newblock {\em The Computer Journal}, 6(3):232--236, 11 1963.

\bibitem{abs-2212-11151}
Yutaka Nagashima, Zijin Xu, Ningli Wang, Daniel~Sebastian Goc, and James Bang.
\newblock Property-based conjecturing for automated induction in isabelle/hol.
\newblock {\em CoRR}, abs/2212.11151, 2022.

\bibitem{abs-2210-03590}
Jelle Piepenbrock, Josef Urban, Konstantin Korovin, Miroslav Ols{\'{a}}k, Tom
  Heskes, and Mikolas Janota.
\newblock Machine learning meets the {H}erbrand {U}niverse.
\newblock {\em CoRR}, abs/2210.03590, 2022.

\bibitem{PiotrowskiU18}
Bartosz Piotrowski and Josef Urban.
\newblock {ATPboost}: Learning premise selection in binary setting with {ATP}
  feedback.
\newblock In Didier Galmiche, Stephan Schulz, and Roberto Sebastiani, editors,
  {\em Automated Reasoning - 9th International Joint Conference, {IJCAR} 2018,
  Held as Part of the Federated Logic Conference, FloC 2018, Oxford, UK, July
  14-17, 2018, Proceedings}, volume 10900 of {\em Lecture Notes in Computer
  Science}, pages 566--574. Springer, 2018.

\bibitem{abs-1911-04873}
Bartosz Piotrowski, Josef Urban, Chad~E. Brown, and Cezary Kaliszyk.
\newblock Can neural networks learn symbolic rewriting?
\newblock {\em CoRR}, abs/1911.04873, 2019.

\bibitem{polya1990mathematics}
George P{\'o}lya.
\newblock {\em Mathematics and plausible reasoning: Induction and analogy in
  mathematics}, volume~1.
\newblock Princeton University Press, 1990.

\bibitem{ReynoldsBNBT19}
Andrew Reynolds, Haniel Barbosa, Andres N{\"{o}}tzli, Clark~W. Barrett, and
  Cesare Tinelli.
\newblock cvc4sy: Smart and fast term enumeration for syntax-guided synthesis.
\newblock In {\em {CAV} {(2)}}, volume 11562 of {\em Lecture Notes in Computer
  Science}, pages 74--83. Springer, 2019.

\bibitem{SchaferS15}
Simon Sch{\"{a}}fer and Stephan Schulz.
\newblock Breeding theorem proving heuristics with genetic algorithms.
\newblock In Gottlob et~al. \cite{DBLP:conf/gcai/2015}, pages 263--274.

\bibitem{silver2017mastering}
David Silver, Julian Schrittwieser, Karen Simonyan, Ioannis Antonoglou, Aja
  Huang, Arthur Guez, Thomas Hubert, Lucas Baker, Matthew Lai, Adrian Bolton,
  Yutian Chen, Timothy Lillicrap, Fan Hui, Laurent Sifre, George van~den
  Driessche, Thore Graepel, and Demis Hassabis.
\newblock Mastering the game of {G}o without human knowledge.
\newblock {\em Nature}, 550:354--, 2017.

\bibitem{oeis}
Neil J.~A. Sloane.
\newblock The {O}n-{L}ine {E}ncyclopedia of {I}nteger {S}equences.
\newblock In Manuel Kauers, Manfred Kerber, Robert Miner, and Wolfgang
  Windsteiger, editors, {\em Towards Mechanized Mathematical Assistants, 14th
  Symposium, Calculemus 2007, 6th International Conference, {MKM} 2007,
  Hagenberg, Austria, June 27-30, 2007, Proceedings}, volume 4573 of {\em
  Lecture Notes in Computer Science}, page 130. Springer, 2007.

\bibitem{SOLOMONOFF19641}
R.J. Solomonoff.
\newblock A formal theory of inductive inference. {P}art {I}.
\newblock {\em Information and Control}, 7(1):1--22, 1964.

\bibitem{Tammet98}
Tanel Tammet.
\newblock Towards efficient subsumption.
\newblock In {\em {CADE}}, volume 1421 of {\em Lecture Notes in Computer
  Science}, pages 427--441. Springer, 1998.

\bibitem{turing1950computing}
A.~M. Turing.
\newblock Computing machinery and intelligence.
\newblock {\em Mind}, 59(236):433--460, 1950.

\bibitem{Urban07}
Josef Urban.
\newblock {MaLARea}: a metasystem for automated reasoning in large theories.
\newblock In Geoff Sutcliffe, Josef Urban, and Stephan Schulz, editors, {\em
  ESARLT}, volume 257 of {\em CEUR Workshop Proceedings}. CEUR-WS.org, 2007.

\bibitem{blistr}
Josef Urban.
\newblock {BliStr: The Blind Strategymaker}.
\newblock In Gottlob et~al. \cite{DBLP:conf/gcai/2015}, pages 312--319.

\bibitem{DBLP:conf/mkm/UrbanJ20}
Josef Urban and Jan Jakubuv.
\newblock First neural conjecturing datasets and experiments.
\newblock In {\em {CICM}}, volume 12236 of {\em Lecture Notes in Computer
  Science}, pages 315--323. Springer, 2020.

\bibitem{US+08}
Josef Urban, Geoff Sutcliffe, Petr Pudl{\'a}k, and Ji\v{r}\'{\i} Vysko\v{c}il.
\newblock {MaLARea SG1 - Machine Learner for Automated Reasoning with Semantic
  Guidance}.
\newblock In Alessandro Armando, Peter Baumgartner, and Gilles Dowek, editors,
  {\em International Joint Conference on Automated Reasoning (IJCAR)}, volume
  5195 of {\em LNCS}, pages 441--456. Springer, 2008.

\bibitem{DBLP:conf/nips/VaswaniSPUJGKP17}
Ashish Vaswani, Noam Shazeer, Niki Parmar, Jakob Uszkoreit, Llion Jones,
  Aidan~N. Gomez, Lukasz Kaiser, and Illia Polosukhin.
\newblock Attention is all you need.
\newblock In Isabelle Guyon, Ulrike von Luxburg, Samy Bengio, Hanna~M. Wallach,
  Rob Fergus, S.~V.~N. Vishwanathan, and Roman Garnett, editors, {\em Advances
  in Neural Information Processing Systems 30: Annual Conference on Neural
  Information Processing Systems 2017, 4-9 December 2017, Long Beach, CA,
  {USA}}, pages 5998--6008, 2017.

\bibitem{WangKU18}
Qingxiang Wang, Cezary Kaliszyk, and Josef Urban.
\newblock First experiments with neural translation of informal to formal
  mathematics.
\newblock In {\em {CICM}}, volume 11006 of {\em Lecture Notes in Computer
  Science}, pages 255--270. Springer, 2018.

\bibitem{DBLP:journals/jintseq/WangZ15}
Yi~Wang and Zhi{-}Hai Zhang.
\newblock Combinatorics of generalized motzkin numbers.
\newblock {\em J. Integer Seq.}, 18(2):15.2.4, 2015.

\end{thebibliography}

\newpage
\appendix

\section{Resources Used by the Experiment}

The average GPU power consumption during inference is 800W: 200W per
each of the four GTX 1080 Ti cards used for inference.
The average GPU power consumption during training is 350W: 175W per
each of the two GTX 1080 Ti cards used for the two kinds of training done in most of the iterations.

The training takes approximately 3 hours and the inference 10.5 hours
(iteration 140 of the longest run). This means that the average GPU consumption per hour
is $(3*350+10.5*800)/13.5 = 700W$.
Additionally, we estimate the non-GPU (CPUs, disks, etc.) consumption
to be on average 150W per hour (the server's CPUs are mostly
idle). %
It took about three months to do the 190 iterations of the longest run. This is $90*24=2160$ hours.
The estimated power consumption for the 190 iterations is thus $2160*(700+150) = 1836$ kWh. 
Assuming a cost in the range of 150-250 EUR per MWh, the main experiment's electricity cost is between 270 and 460 EUR.
The three shorter experiments have so far run for about half as many iterations and are also using on average less resources (fewer GPUs, shorter inference length). The estimated power consumption for all experiments is thus likely below 4 MWh, and the total electricity cost below 1000 EUR. %

\section{Triangle Coding}\label{sec:triangle}
Our system has found a correspondence between a pair of non-negative integers $(x_a,x_b)$ where $x_a \geq x_b$, and a single
non-negative integer $x$ by enumerating the following sequence of pairs $(x_a,x_b)$ with non-negative integers:
$$(0,0), (1,0), (1,1), (2,0), (2,1), (2,2), (3,0), \ldots.$$
In one direction, it computes $x = \frac{x_a \times (x_a + 1)}{2} + x_b$ in order to ``encode'' the pair $(x_a, x_b)$. Decoding
of a single non-negative integer $x$ can be calculated as $(x_a, x_b) = (f_0(x)-f_1(x), f_0(x))$, where the functions $f_0, f_1$ can be implemented in Python for example as follows (an actual code invented by the program generator). Note that the function $f_0$ computes $x_b$, and the function $f_1$ computes $x_b - x_a$ (a non-positive integer).

\begin{lstlisting}[language=Python]
def f0(X):
    x = X
    for y in range (1,(2 + (X // (1 + (2 + 2)))) + 1):
        x = x - (y if (y - x) <= 0 else 0)
    return x

def f1(X):
    x = X
    for y in range (1,(2 + (X // (1 + (2 + 2)))) + 1):
        x = x - (0 if x <= 0 else (1 + y))
    return x
\end{lstlisting}

This representation was initially discovered when our system found solutions
for the  sequences A2262,\footnoteA{\url{https://oeis.org/A002262}} and
A25581.\footnoteA{\url{https://oeis.org/A025581}} Indeed, $f_0$ is a solution for
A2262 invented in the first iteration and $-f_1$ is a solution for A25581
invented during the $9^{th}$ iteration.

\subsection{Number of steps}

\begin{figure}
  \centering
  \includegraphics[width=12cm]{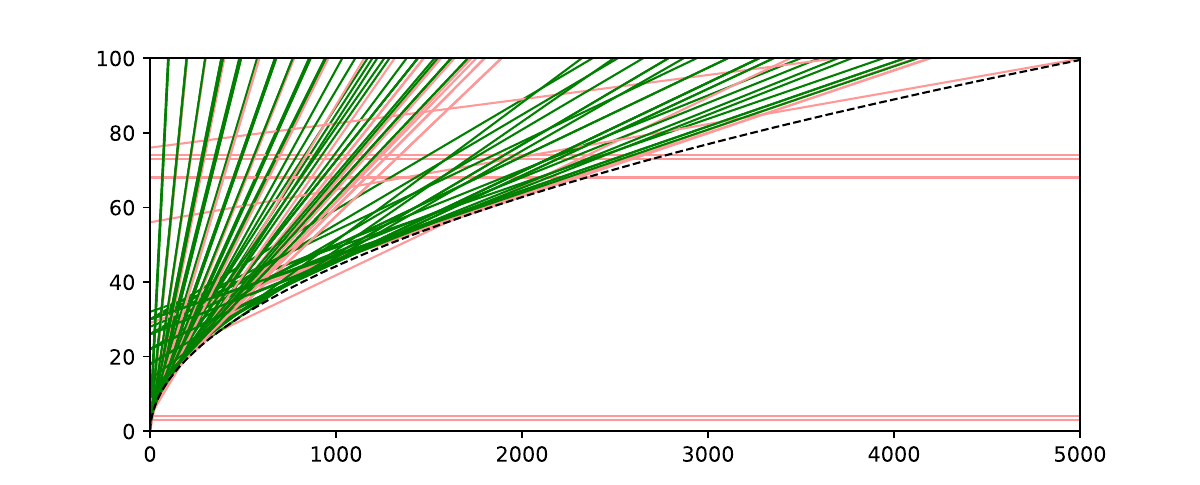}
  \caption{Linear bounds (in red or green) for
  $y=\sqrt{2x+\frac14}-\frac12$ 
  (in black) invented by the 
  system with $x \leq 5000$.
  }
  \label{fig:sqrt1}
\end{figure}

\begin{figure}
  \centering
  \includegraphics[width=12cm]{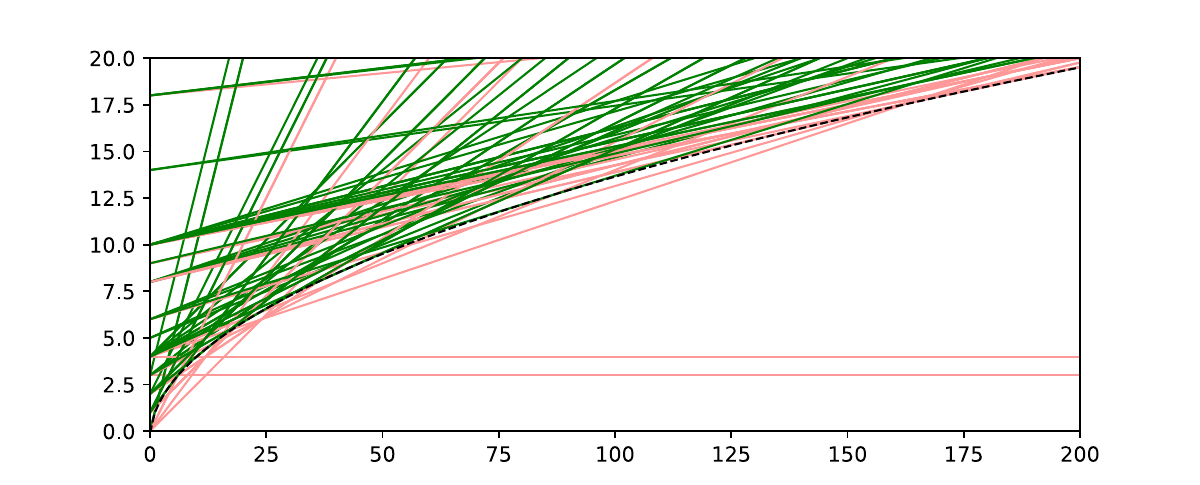}
  \caption{Linear bounds (in red or green) for
  $y=\sqrt{2x+\frac14}-\frac12$ 
  (in black) invented by the 
  system with $x \leq 200$.}
  \label{fig:sqrt2}
\end{figure}

In order to calculate $f_0$ and $f_1$, one must perform approximately
$\sqrt{2x+\frac14}-\frac12$ steps in a subtracting loop. Since the
programming language is based on \texttt{for} loops, the programs are
approximating the number of subtracting steps with a function that
can be easily obtained using the basic arithmetical operations. In
particular, the code above uses approximations with $2+x/5$ where 5 is
written as $1+(2+2)$ (our basic language only supports constants
0,1,2).  The program generator has experimented with many (over 80) such bounds,
the vast majority of such bounds are of the shape $a+x/b$ or $a+2(x/b)$
where $a$, $b$ are constants.  As the speed of computing with larger numbers becomes more important, the slope of the approximation flattens, with $b$ reaching values over $50$. We have collected all such subprograms
used in the generated programs, and plotted the bounds in
Figure~\ref{fig:sqrt1} and Figure~\ref{fig:sqrt2}. The approximated
function $\sqrt{2x+\frac14}-\frac12$ is plotted with a black dashed
line. All the found valid bounds are plotted in green, all the invalid
bound attempts are plotted in red. 

\subsection{Triangle coding in action}

The triangle coding turned out to be useful in many cases.
The designed language for our programs only supports a maximum of two variables but using the triangle coding,
a pair of variables can be temporarily packed into one. A simple example is sequence
A279364\footnoteA{\url{https://oeis.org/A279364}} -- sum of 5th
powers of proper divisors of n. A human programmer could implement this sequence as:

\begin{lstlisting}[language=Python]
def f(X):
    res = 0
    for y in range(1,X+1):
        res = res + (y**5 if (X+1) %
    return res
\end{lstlisting}

There are three variables used in the body of the loop, in
particular \texttt{res}, \texttt{y}, and \texttt{X}. Since the native language
supports only two variables available at every moment, this Python code cannot
be straightforwardly translated into the language of our programs. Nevertheless,
the program generator has found a workaround using
the triangle coding -- it packs \texttt{X} and \texttt{y} into a single natural
number, so it can perform the summing loop adding to \texttt{res}, and in order
to decode what to add to \texttt{res}, it unpacks the pair, and checks whether
\texttt{y} divides \texttt{X+1}.

\paragraph{The code generated by NMT for A279364} is as follows:
\begin{center}
\fontsize{8}{10}\selectfont
\begin{verbatim}
A279364 Sum of 5th powers of proper divisors of n.
371326 59293 555688 1 870552 1 1082401 161295 1419890 19933 2206525 1
2476132 371537 3336950 1 4646784 1 5315740 821793 6436376 1 9301876
16808 9868783

size 94, time 1335684: loop2 ((loop (loop2 ((loop ((((x * x) * x) * x)
* x) 1 (1 + y)) * (if (x mod (1 + y)) <= 0 then 1 else 0)) 0 1 (1 -
(loop (x - (if x <= 0 then 0 else y)) (1 + (2 + (2 + (x div (1 + (2 *
(2 + 2))))))) (1 + x))) (loop (x - (if (y - x) <= 0 then y else 0)) (2
+ (2 + (x div (1 + (2 * (2 + 2)))))) x)) 1 y) + x) (1 + y) x 0 (((x *
x) - x) div 2)

K K F K F K F K F B B L D J K B L D H B A I F A B B K K A L I E B C C
K B C C C D F D G D D D B K D J E K L K E L A I E C C K B C C C D F D
G D D K J N B L J K D B L D K A K K F K E C G N
\end{verbatim}
\end{center}  

\begin{lstlisting}[language=Python]
def f3(X,Y):
   x = 1 + Y
   for y in range (1,1 + 1):
       x = (((x * x) * x) * x) * x
   return x

def f4(X):
   x = 1 + X
   for y in range (1,(1 + (2 + (2 + (X // 9)))) + 1):
       x = x - (0 if x <= 0 else y)
   return x

def f5(X):
   x = X
   for y in range (1,(2 + (2 + (X // 9))) + 1):
       x = x - (y if (y - x) <= 0 else 0)
   return x

def f2(X):
   x,y = 1 - f4(X), f5(X)
   for z in range (1,1 + 1):
       x,y = (f3(x,y) * (1 if (x %
   return x

def f1(X,Y):
   x = Y
   for y in range (1,1 + 1):
       x = f2(x)
   return x

def f0(X):
   x,y = 0, ((X * X) - X) // 2
   for z in range (1,X + 1):
       x,y = (f1(x,y) + x), (1 + y)
   return x

for x in range(32):
   print (f0(x))
\end{lstlisting}

The number $9$ in these functions is used as a shorthand for $1+2*(2+2)$.
Functions \texttt{f4}, \texttt{f5} are calculating the triangle coding, \texttt{f3} is the fifth power, \texttt{f1} is just a dummy function using \texttt{Y} instead of \texttt{X} for \texttt{f2}.
Finally, \texttt{f2} is returning either $(b+1)^5$, or 0 depending on whether $b+1$ divides $a+2$ or not where $(a,b)$ if the triangle coding pair of \texttt{X}, and \texttt{f0} is summing over
all the pairs $(a,b)$ going from \texttt{(X-1,0)} to \texttt{(X-1,X-1)}. 

\section{Evolution and Proliferation}

We analyze the evolution of the solutions found for the OEIS sequence of
prime numbers.\footnoteA{The tables shown here for primes are also in our repository~\url{https://bit.ly/3XHZsjK}, together with similar tables for the sigma function (sum of the divisors).} The exact iterations of their discovery are shown in
our repository, together with their size and
speed.\footnoteA{\url{https://bit.ly/3iJ4oGd}} The 24 invented programs are shown in Table~\ref{tab:primes1}.

\begin{table}
  \caption{The 24 programs for primes.}\label{tab:primes1}
          \begin{center}
			\resizebox{\textwidth}{!}{%
				\begin{tabular}{l|p{13cm}}
                          \toprule
                          Nr & Program \\
                          \midrule
P1 &  (if x <= 0 then 2 else 1) + (compr (((loop (x + x) (x mod 2) (loop (x * x) 1 (loop (x + x) (x div 2) 1))) + x) mod (1 + x)) x)\\
P2 &  1 + (compr ((((loop (x * x) 1 (loop (x + x) (x div 2) 1)) + x) * x) mod (1 + x)) (1 + x))\\
P3 &  1 + (compr (((loop (x * x) 1 (loop (x + x) (x div 2) 1)) + x) mod (1 + x)) (1 + x))\\
P4 &  2 + (compr ((loop2 (1 + (if (x mod (1 + y)) <= 0 then 0 else x)) (y - 1) x 1 x) mod (1 + x)) x)\\
P5 &  1 + (compr ((loop (if (x mod (1 + y)) <= 0 then (1 + y) else x) x (1 + x)) mod (1 + x)) (1 + x))\\
P6 &  1 + (compr ((loop (if (x mod (1 + y)) <= 0 then (1 + y) else x) (2 + (x div (2 + (2 + 2)))) (1 + x)) mod (1 + x)) (1 + x))\\
P7 &  compr ((1 + (loop (if (x mod (1 + y)) <= 0 then (1 + y) else x) x x)) mod (1 + x)) (2 + x)\\
P8 &  1 + (compr ((loop (if (x mod (1 + y)) <= 0 then (1 + y) else x) (1 + ((2 + x) div (2 + (2 + 2)))) (1 + x)) mod (1 + x)) (1 + x))\\
P9 &  compr (x - (loop (if (x mod (1 + y)) <= 0 then (1 + y) else x) x x)) (2 + x)\\
P10 & compr (x - (loop (if (x mod (1 + y)) <= 0 then 2 else x) (x div 2) x)) (2 + x)\\
P11 & 1 + (compr ((loop (if (x mod (1 + y)) <= 0 then (1 + y) else x) (1 + (x div (2 + (2 + 2)))) (1 + x)) mod (1 + x)) (1 + x))\\
P12 & compr ((x - (loop (if (x mod (1 + y)) <= 0 then y else x) x x)) - 2) (2 + x)\\
P13 & 1 + (compr ((loop (if (x mod (1 + y)) <= 0 then (1 + y) else x) (2 + (x div (2 * (2 + (2 + 2))))) (1 + x)) mod (1 + x)) (1 + x))\\
P14 & compr ((x - (loop (if (x mod (1 + y)) <= 0 then y else x) x x)) - 1) (2 + x)\\
P15 & 1 + (compr (x - (loop (if (x mod (1 + y)) <= 0 then (1 + y) else x) (2 + (x div (2 * (2 + (2 + 2))))) (1 + x))) (1 + x))\\
P16 & compr (2 - (loop (if (x mod (1 + y)) <= 0 then 0 else x) (x - 2) x)) x\\
P17 & 1 + (compr (x - (loop (if (x mod (1 + y)) <= 0 then 2 else x) (2 + (x div (2 * (2 + (2 + 2))))) (1 + x))) (1 + x))\\
P18 & 1 + (compr (x - (loop (if (x mod (1 + y)) <= 0 then 2 else x) (1 + (2 + (x div (2 * (2 * (2 + 2)))))) (1 + x))) (1 + x))\\
P19 & 1 + (compr (x - (loop2 (loop (if (x mod (1 + y)) <= 0 then 2 else x) (2 + (y div (2 * (2 + (2 + 2))))) (1 + y)) 0 (1 - (x mod 2)) 1 x)) (1 + x))\\
P20 & 1 + (compr (x - (loop2 (loop (if (x mod (1 + y)) <= 0 then 2 else x) (1 + (2 + (y div (2 * (2 * (2 + 2)))))) (1 + y)) 0 (1 - (x mod 2)) 1 x)) (1 + x))\\
P21 & 1 + (compr (x - (loop2 (loop (if (x mod (2 + y)) <= 0 then 2 else x) (2 + (y div (2 * ((2 + 2) + (2 + 2))))) (1 + y)) 0 (1 - (x mod 2)) 1 x)) (1 + x))\\
P22 & 1 + (compr (x - (loop2 (loop (if (x mod (2 + y)) <= 0 then 2 else x) (2 + (y div (2 * (2 * (2 + 2))))) (1 + y)) 0 (1 - (x mod 2)) 1 x)) (1 + x))\\
P23 & 2 + (compr (loop (x - (if (x mod (1 + y)) <= 0 then 0 else 1)) x x) x)\\
P24 & loop (1 + x) (1 - x) (1 + (2 * (compr (x - (loop (if (x mod (2 + y)) <= 0 then 1 else x) (2 + (x div (2 * (2 + 2)))) (1 + (x + x)))) x)))\\
\bottomrule
\end{tabular}}
\end{center}
\end{table}

We can observe how these 24 programs propagate through the population
of all programs, as they evolve in the iterations. This is shown in
Table~\ref{tab:prolifp1} and Table~\ref{tab:prolifp2}. We can see
that, e.g., P5 (size 25, time 515990), gets quickly replaced by P6
(size 33, time 98390) and P7 (size 23, time 519654) after their
invention. P6 is more than five times faster than P5, while P7 is
smaller. Therefore they are included in the training data instead of
P5 when they are invented. While there are still other programs in the
training data using P5 as a subprogram, their faster/smaller versions
get eventually also reinvented as the newly trained NMT increasingly
prefers to synthesize programs that contain P6 or P7. This illustrates
the dynamics of the overall alien system. The training of the neural
synthesis component (NMT) evolves, governed by more high-level
evolutionary fitness criteria, which are in our case size (Occam's
razor) and speed (efficiency).

	\begin{table}[hbt!]
  \caption{Proliferation of the 24 programs for primes.}\label{tab:prolifp1}
        \begin{center}
		\resizebox{\textwidth}{!}{%
				\begin{tabular}{l|llllllllllllllllllllllll}
                          \toprule
                          Iter
                          &P1&P2&P3&P4&P5&P6&P7&P8&P9&P10&P11&P12&P13&P14&P15&P16&P17&P18&P19&P20&P21&P22&P23&P24\\
                          \midrule
25&0 & 0 & 0 & 0 & 0 & 0 & 0 & 0 & 0 & 0 & 0 & 0 & 0 & 0 & 0 & 0 & 0 & 0 & 0 & 0 & 0 & 0 & 0 & 0\\
26&6 & 0 & 0 & 0 & 0 & 0 & 0 & 0 & 0 & 0 & 0 & 0 & 0 & 0 & 0 & 0 & 0 & 0 & 0 & 0 & 0 & 0 & 0 & 0\\
27&7 & 0 & 0 & 0 & 0 & 0 & 0 & 0 & 0 & 0 & 0 & 0 & 0 & 0 & 0 & 0 & 0 & 0 & 0 & 0 & 0 & 0 & 0 & 0\\
28&8 & 0 & 0 & 0 & 0 & 0 & 0 & 0 & 0 & 0 & 0 & 0 & 0 & 0 & 0 & 0 & 0 & 0 & 0 & 0 & 0 & 0 & 0 & 0\\
29&9 & 0 & 0 & 0 & 0 & 0 & 0 & 0 & 0 & 0 & 0 & 0 & 0 & 0 & 0 & 0 & 0 & 0 & 0 & 0 & 0 & 0 & 0 & 0\\
30&10 & 0 & 0 & 0 & 0 & 0 & 0 & 0 & 0 & 0 & 0 & 0 & 0 & 0 & 0 & 0 & 0 & 0 & 0 & 0 & 0 & 0 & 0 & 0\\
31&4 & 6 & 0 & 0 & 0 & 0 & 0 & 0 & 0 & 0 & 0 & 0 & 0 & 0 & 0 & 0 & 0 & 0 & 0 & 0 & 0 & 0 & 0 & 0\\
32&6 & 6 & 0 & 0 & 0 & 0 & 0 & 0 & 0 & 0 & 0 & 0 & 0 & 0 & 0 & 0 & 0 & 0 & 0 & 0 & 0 & 0 & 0 & 0\\
33&8 & 1 & 6 & 6 & 0 & 0 & 0 & 0 & 0 & 0 & 0 & 0 & 0 & 0 & 0 & 0 & 0 & 0 & 0 & 0 & 0 & 0 & 0 & 0\\
34&12 & 4 & 6 & 6 & 1 & 0 & 0 & 0 & 0 & 0 & 0 & 0 & 0 & 0 & 0 & 0 & 0 & 0 & 0 & 0 & 0 & 0 & 0 & 0\\
35&7 & 12 & 6 & 0 & 9 & 0 & 0 & 0 & 0 & 0 & 0 & 0 & 0 & 0 & 0 & 0 & 0 & 0 & 0 & 0 & 0 & 0 & 0 & 0\\
36&4 & 10 & 6 & 0 & 17 & 0 & 0 & 0 & 0 & 0 & 0 & 0 & 0 & 0 & 0 & 0 & 0 & 0 & 0 & 0 & 0 & 0 & 0 & 0\\
37&3 & 4 & 6 & 0 & 18 & 6 & 1 & 0 & 0 & 0 & 0 & 0 & 0 & 0 & 0 & 0 & 0 & 0 & 0 & 0 & 0 & 0 & 0 & 0\\
38&2 & 3 & 1 & 0 & 12 & 18 & 11 & 0 & 0 & 0 & 0 & 0 & 0 & 0 & 0 & 0 & 0 & 0 & 0 & 0 & 0 & 0 & 0 & 0\\
39&2 & 3 & 1 & 0 & 9 & 56 & 31 & 0 & 0 & 0 & 0 & 0 & 0 & 0 & 0 & 0 & 0 & 0 & 0 & 0 & 0 & 0 & 0 & 0\\
40&2 & 5 & 2 & 0 & 7 & 59 & 49 & 9 & 1 & 0 & 2 & 0 & 1 & 0 & 0 & 0 & 0 & 0 & 0 & 0 & 0 & 0 & 0 & 0\\
41&1 & 2 & 3 & 0 & 4 & 52 & 58 & 42 & 23 & 0 & 13 & 0 & 8 & 0 & 0 & 0 & 0 & 0 & 0 & 0 & 0 & 0 & 0 & 0\\
42&0 & 2 & 4 & 0 & 3 & 44 & 50 & 38 & 60 & 8 & 11 & 0 & 55 & 0 & 0 & 0 & 0 & 0 & 0 & 0 & 0 & 0 & 0 & 0\\
43&0 & 2 & 12 & 0 & 0 & 37 & 55 & 14 & 116 & 35 & 16 & 7 & 90 & 0 & 0 & 0 & 0 & 0 & 0 & 0 & 0 & 0 & 0 & 0\\
44&0 & 2 & 13 & 0 & 0 & 28 & 40 & 6 & 176 & 73 & 19 & 8 & 122 & 9 & 12 & 0 & 0 & 0 & 0 & 0 & 0 & 0 & 0 & 0\\
45&0 & 2 & 9 & 0 & 0 & 19 & 24 & 4 & 147 & 185 & 26 & 16 & 94 & 25 & 29 & 0 & 7 & 0 & 0 & 0 & 0 & 0 & 0 & 0\\
46&0 & 2 & 4 & 0 & 0 & 11 & 14 & 0 & 101 & 256 & 21 & 14 & 66 & 64 & 30 & 0 & 29 & 0 & 0 & 0 & 0 & 0 & 0 & 0\\
47&0 & 0 & 0 & 0 & 0 & 9 & 4 & 0 & 55 & 290 & 23 & 3 & 43 & 116 & 16 & 6 & 62 & 14 & 0 & 0 & 0 & 0 & 0 & 0\\
48&0 & 0 & 0 & 0 & 0 & 8 & 0 & 0 & 22 & 261 & 16 & 0 & 34 & 192 & 10 & 6 & 89 & 30 & 0 & 0 & 0 & 0 & 0 & 0\\
49&0 & 0 & 0 & 0 & 0 & 8 & 0 & 0 & 6 & 195 & 11 & 0 & 36 & 225 & 8 & 6 & 99 & 34 & 0 & 0 & 0 & 0 & 0 & 0\\
50&0 & 0 & 0 & 0 & 0 & 5 & 0 & 0 & 2 & 154 & 8 & 0 & 29 & 168 & 6 & 6 & 108 & 39 & 0 & 0 & 0 & 0 & 0 & 0\\
51&0 & 0 & 0 & 0 & 0 & 4 & 0 & 0 & 0 & 121 & 7 & 0 & 21 & 97 & 6 & 6 & 113 & 43 & 0 & 0 & 0 & 0 & 0 & 0\\
52&0 & 0 & 0 & 0 & 0 & 2 & 0 & 0 & 0 & 118 & 8 & 0 & 12 & 62 & 6 & 6 & 110 & 51 & 0 & 0 & 0 & 0 & 0 & 0\\
53&0 & 0 & 0 & 0 & 0 & 1 & 0 & 0 & 0 & 59 & 7 & 0 & 15 & 33 & 6 & 6 & 125 & 62 & 0 & 0 & 0 & 0 & 0 & 0\\
54&0 & 0 & 0 & 0 & 0 & 1 & 0 & 0 & 0 & 41 & 4 & 0 & 16 & 17 & 6 & 9 & 137 & 72 & 0 & 0 & 0 & 0 & 0 & 0\\
55&0 & 0 & 0 & 0 & 0 & 2 & 0 & 0 & 0 & 32 & 4 & 0 & 15 & 9 & 6 & 17 & 147 & 82 & 0 & 0 & 0 & 0 & 0 & 0\\
56&0 & 0 & 0 & 0 & 0 & 1 & 0 & 0 & 0 & 29 & 4 & 0 & 10 & 7 & 6 & 39 & 152 & 98 & 0 & 0 & 0 & 0 & 0 & 0\\
57&0 & 0 & 0 & 0 & 0 & 1 & 0 & 0 & 1 & 28 & 3 & 0 & 9 & 5 & 6 & 103 & 142 & 108 & 0 & 0 & 0 & 0 & 0 & 0\\
58&0 & 0 & 0 & 0 & 0 & 0 & 0 & 0 & 1 & 17 & 3 & 0 & 7 & 4 & 6 & 146 & 146 & 120 & 0 & 0 & 0 & 0 & 0 & 0\\
59&0 & 0 & 0 & 0 & 0 & 0 & 0 & 0 & 1 & 11 & 3 & 0 & 6 & 2 & 0 & 179 & 153 & 122 & 0 & 0 & 0 & 0 & 0 & 0\\
60&0 & 0 & 0 & 0 & 0 & 0 & 0 & 0 & 1 & 6 & 3 & 0 & 3 & 2 & 0 & 206 & 148 & 121 & 0 & 0 & 0 & 0 & 0 & 0\\
61&0 & 0 & 0 & 0 & 0 & 0 & 0 & 0 & 0 & 6 & 3 & 0 & 3 & 1 & 0 & 220 & 139 & 138 & 0 & 0 & 0 & 0 & 0 & 0\\
62&0 & 0 & 0 & 0 & 0 & 0 & 0 & 0 & 0 & 5 & 3 & 0 & 2 & 0 & 0 & 245 & 118 & 145 & 0 & 0 & 0 & 0 & 0 & 0\\
63&0 & 0 & 0 & 0 & 0 & 0 & 0 & 0 & 0 & 5 & 3 & 0 & 2 & 0 & 0 & 263 & 103 & 160 & 0 & 0 & 0 & 0 & 0 & 0\\
64&0 & 0 & 3 & 0 & 0 & 0 & 0 & 0 & 0 & 6 & 4 & 0 & 2 & 0 & 0 & 284 & 87 & 162 & 0 & 0 & 0 & 0 & 0 & 0\\
65&0 & 0 & 13 & 0 & 0 & 0 & 0 & 0 & 0 & 5 & 4 & 0 & 1 & 0 & 0 & 303 & 74 & 173 & 0 & 0 & 0 & 0 & 0 & 0\\
66&0 & 0 & 34 & 0 & 0 & 0 & 0 & 0 & 0 & 3 & 2 & 0 & 1 & 0 & 0 & 315 & 67 & 174 & 0 & 0 & 0 & 0 & 0 & 0\\
67&0 & 0 & 53 & 0 & 0 & 0 & 0 & 0 & 0 & 1 & 1 & 0 & 1 & 0 & 0 & 321 & 64 & 180 & 0 & 0 & 0 & 0 & 0 & 0\\
68&0 & 0 & 61 & 0 & 0 & 0 & 0 & 0 & 0 & 1 & 1 & 0 & 1 & 0 & 0 & 323 & 66 & 178 & 0 & 0 & 0 & 0 & 0 & 0\\
69&0 & 0 & 67 & 0 & 0 & 0 & 0 & 0 & 0 & 1 & 1 & 0 & 1 & 0 & 0 & 325 & 63 & 178 & 0 & 0 & 0 & 0 & 0 & 0\\
70&0 & 0 & 70 & 0 & 0 & 0 & 0 & 0 & 0 & 1 & 1 & 0 & 1 & 0 & 0 & 324 & 60 & 182 & 0 & 0 & 0 & 0 & 0 & 0\\
71&0 & 0 & 72 & 0 & 0 & 0 & 0 & 0 & 0 & 1 & 1 & 0 & 1 & 0 & 0 & 330 & 56 & 181 & 0 & 0 & 0 & 0 & 0 & 0\\
72&0 & 0 & 73 & 0 & 0 & 0 & 0 & 0 & 0 & 0 & 2 & 0 & 1 & 0 & 0 & 332 & 57 & 190 & 0 & 0 & 0 & 0 & 0 & 0\\
73&0 & 0 & 72 & 0 & 0 & 0 & 0 & 0 & 0 & 0 & 2 & 0 & 1 & 0 & 0 & 330 & 58 & 191 & 0 & 0 & 0 & 0 & 0 & 0\\
74&0 & 0 & 71 & 0 & 0 & 0 & 0 & 0 & 0 & 0 & 3 & 0 & 1 & 0 & 0 & 336 & 56 & 189 & 0 & 0 & 0 & 0 & 0 & 0\\
75&0 & 0 & 74 & 0 & 0 & 0 & 0 & 0 & 0 & 0 & 2 & 0 & 1 & 0 & 0 & 340 & 55 & 192 & 0 & 0 & 0 & 0 & 0 & 0\\
76&0 & 0 & 77 & 0 & 0 & 0 & 0 & 0 & 0 & 0 & 1 & 0 & 0 & 0 & 0 & 341 & 57 & 195 & 0 & 0 & 0 & 0 & 0 & 0\\
77&0 & 0 & 79 & 0 & 0 & 0 & 0 & 0 & 0 & 0 & 1 & 0 & 0 & 0 & 0 & 343 & 56 & 191 & 0 & 0 & 0 & 0 & 0 & 0\\
78&0 & 0 & 79 & 0 & 0 & 0 & 0 & 0 & 0 & 0 & 1 & 0 & 0 & 0 & 0 & 344 & 57 & 201 & 0 & 0 & 0 & 0 & 0 & 0\\
79&0 & 0 & 81 & 0 & 0 & 0 & 0 & 0 & 0 & 0 & 0 & 0 & 0 & 0 & 0 & 344 & 56 & 200 & 0 & 0 & 0 & 0 & 0 & 0\\
80&0 & 0 & 80 & 0 & 0 & 0 & 0 & 0 & 0 & 0 & 0 & 0 & 0 & 0 & 0 & 346 & 55 & 210 & 0 & 0 & 0 & 0 & 0 & 0\\
81&0 & 0 & 75 & 0 & 0 & 0 & 0 & 0 & 0 & 0 & 0 & 0 & 0 & 0 & 0 & 351 & 55 & 206 & 0 & 0 & 0 & 0 & 0 & 0\\
82&0 & 0 & 77 & 0 & 0 & 0 & 0 & 0 & 0 & 0 & 0 & 0 & 0 & 0 & 0 & 354 & 53 & 206 & 0 & 0 & 0 & 0 & 0 & 0\\
83&0 & 0 & 77 & 0 & 0 & 0 & 0 & 0 & 0 & 1 & 0 & 0 & 0 & 0 & 0 & 360 & 53 & 207 & 0 & 0 & 0 & 0 & 0 & 0\\
84&0 & 0 & 76 & 0 & 0 & 0 & 0 & 0 & 0 & 1 & 0 & 0 & 0 & 0 & 0 & 360 & 53 & 208 & 0 & 0 & 0 & 0 & 0 & 0\\
85&0 & 0 & 74 & 0 & 0 & 0 & 0 & 0 & 0 & 1 & 0 & 0 & 0 & 0 & 0 & 363 & 53 & 207 & 0 & 0 & 0 & 0 & 0 & 0\\
86&0 & 0 & 75 & 0 & 0 & 0 & 0 & 0 & 0 & 0 & 0 & 0 & 0 & 0 & 0 & 361 & 54 & 205 & 0 & 0 & 0 & 0 & 0 & 0\\
87&0 & 0 & 77 & 0 & 0 & 0 & 0 & 0 & 0 & 0 & 0 & 0 & 0 & 0 & 0 & 359 & 54 & 198 & 0 & 0 & 0 & 0 & 0 & 0\\
88&0 & 0 & 80 & 0 & 0 & 0 & 0 & 0 & 0 & 1 & 0 & 0 & 0 & 0 & 0 & 359 & 54 & 199 & 0 & 0 & 0 & 0 & 0 & 0\\
89&0 & 0 & 83 & 0 & 0 & 0 & 0 & 0 & 0 & 1 & 0 & 0 & 0 & 0 & 0 & 357 & 56 & 196 & 0 & 0 & 0 & 0 & 0 & 0\\
                          \bottomrule
                        \end{tabular}%
                    }
                    \end{center}
                  \end{table}
                  
\begin{table}[hbt!]
          \caption{Proliferation of the 24 programs for primes.}\label{tab:prolifp2}
          \begin{center}
            \resizebox{\textwidth}{!}{%
                  \begin{tabular}{l|llllllllllllllllllllllll}
                    \toprule
                    Iter 
                    &P1&P2&P3&P4&P5&P6&P7&P8&P9&P10&P11&P12&P13&P14&P15&P16&P17&P18&P19&P20&P21&P22&P23&P24\\
                    \midrule
90&0 & 0 & 82 & 0 & 0 & 0 & 0 & 0 & 0 & 1 & 0 & 0 & 0 & 0 & 0 & 345 & 56 & 194 & 0 & 0 & 0 & 0 & 0 & 0\\
91&0 & 0 & 81 & 0 & 0 & 0 & 0 & 0 & 0 & 1 & 0 & 0 & 0 & 0 & 0 & 331 & 50 & 188 & 0 & 0 & 0 & 0 & 0 & 0\\
92&0 & 0 & 66 & 0 & 0 & 0 & 0 & 0 & 0 & 0 & 0 & 0 & 0 & 0 & 0 & 322 & 31 & 176 & 9 & 1 & 0 & 0 & 0 & 0\\
93&0 & 0 & 56 & 0 & 0 & 0 & 0 & 0 & 0 & 0 & 0 & 0 & 0 & 0 & 0 & 313 & 18 & 162 & 37 & 17 & 0 & 0 & 0 & 0\\
94&0 & 0 & 41 & 0 & 0 & 0 & 0 & 0 & 0 & 0 & 0 & 0 & 0 & 0 & 0 & 303 & 9 & 145 & 52 & 30 & 0 & 0 & 0 & 0\\
95&0 & 0 & 26 & 0 & 0 & 0 & 0 & 0 & 0 & 0 & 0 & 0 & 0 & 0 & 0 & 300 & 8 & 130 & 54 & 38 & 0 & 0 & 0 & 0\\
96&0 & 0 & 22 & 0 & 0 & 0 & 0 & 0 & 0 & 1 & 0 & 0 & 0 & 0 & 0 & 293 & 5 & 119 & 64 & 47 & 1 & 0 & 0 & 0\\
97&0 & 0 & 17 & 0 & 0 & 0 & 0 & 0 & 0 & 0 & 0 & 0 & 0 & 0 & 0 & 285 & 4 & 100 & 78 & 56 & 2 & 0 & 0 & 0\\
98&0 & 0 & 16 & 0 & 0 & 0 & 0 & 0 & 0 & 0 & 0 & 0 & 0 & 0 & 0 & 276 & 3 & 88 & 79 & 64 & 10 & 0 & 0 & 0\\
99&0 & 0 & 12 & 0 & 0 & 0 & 0 & 0 & 0 & 0 & 0 & 0 & 0 & 0 & 0 & 269 & 0 & 78 & 56 & 56 & 85 & 0 & 0 & 0\\
100&0 & 0 & 10 & 0 & 0 & 0 & 0 & 0 & 0 & 0 & 0 & 0 & 0 & 0 & 0 & 266 & 0 & 73 & 40 & 51 & 116 & 0 & 0 & 0\\
101&0 & 0 & 8 & 0 & 0 & 0 & 0 & 0 & 0 & 0 & 0 & 0 & 0 & 0 & 0 & 266 & 1 & 63 & 33 & 55 & 49 & 0 & 0 & 0\\
102&0 & 0 & 7 & 0 & 0 & 0 & 0 & 0 & 0 & 0 & 0 & 0 & 0 & 0 & 0 & 259 & 0 & 56 & 23 & 64 & 25 & 0 & 0 & 0\\
103&0 & 0 & 5 & 0 & 0 & 0 & 0 & 0 & 0 & 0 & 0 & 0 & 0 & 0 & 0 & 256 & 2 & 50 & 21 & 56 & 29 & 0 & 0 & 0\\
104&0 & 0 & 5 & 0 & 0 & 0 & 0 & 0 & 0 & 0 & 0 & 0 & 0 & 0 & 0 & 258 & 1 & 49 & 20 & 49 & 29 & 0 & 0 & 0\\
105&0 & 0 & 5 & 0 & 0 & 0 & 0 & 0 & 0 & 0 & 0 & 0 & 0 & 0 & 0 & 255 & 1 & 47 & 18 & 43 & 27 & 0 & 0 & 0\\
106&0 & 0 & 5 & 0 & 0 & 0 & 0 & 0 & 0 & 0 & 0 & 0 & 0 & 0 & 0 & 253 & 1 & 44 & 14 & 37 & 15 & 0 & 0 & 0\\
107&0 & 0 & 5 & 0 & 0 & 0 & 0 & 0 & 0 & 0 & 0 & 0 & 0 & 0 & 0 & 252 & 1 & 40 & 12 & 36 & 11 & 0 & 0 & 0\\
108&0 & 0 & 5 & 0 & 0 & 0 & 0 & 0 & 0 & 0 & 0 & 0 & 0 & 0 & 0 & 255 & 2 & 38 & 12 & 34 & 8 & 0 & 0 & 0\\
109&0 & 0 & 4 & 0 & 0 & 0 & 0 & 0 & 0 & 0 & 0 & 0 & 0 & 0 & 0 & 255 & 3 & 34 & 11 & 33 & 8 & 0 & 0 & 0\\
110&0 & 0 & 4 & 0 & 0 & 0 & 0 & 0 & 0 & 0 & 0 & 0 & 0 & 0 & 0 & 256 & 2 & 35 & 10 & 30 & 8 & 0 & 0 & 0\\
111&0 & 0 & 4 & 0 & 0 & 0 & 0 & 0 & 0 & 0 & 0 & 0 & 0 & 0 & 0 & 258 & 2 & 32 & 10 & 31 & 7 & 0 & 0 & 0\\
112&0 & 0 & 4 & 0 & 0 & 0 & 0 & 0 & 0 & 0 & 0 & 0 & 0 & 0 & 0 & 262 & 2 & 31 & 11 & 31 & 7 & 0 & 0 & 0\\
113&0 & 0 & 2 & 0 & 0 & 0 & 0 & 0 & 0 & 0 & 0 & 0 & 0 & 0 & 0 & 263 & 0 & 31 & 10 & 29 & 1 & 0 & 0 & 0\\
114&0 & 0 & 2 & 0 & 0 & 0 & 0 & 0 & 0 & 0 & 0 & 0 & 0 & 0 & 0 & 263 & 0 & 31 & 7 & 30 & 1 & 0 & 0 & 0\\
115&0 & 0 & 1 & 0 & 0 & 0 & 0 & 0 & 0 & 0 & 0 & 0 & 0 & 0 & 0 & 261 & 0 & 30 & 5 & 28 & 1 & 0 & 0 & 0\\
116&0 & 0 & 1 & 0 & 0 & 0 & 0 & 0 & 0 & 0 & 0 & 0 & 0 & 0 & 0 & 263 & 0 & 27 & 6 & 29 & 1 & 0 & 0 & 0\\
117&0 & 0 & 1 & 0 & 0 & 0 & 0 & 0 & 0 & 0 & 0 & 0 & 0 & 0 & 0 & 263 & 0 & 28 & 4 & 27 & 1 & 0 & 0 & 0\\
118&0 & 0 & 1 & 0 & 0 & 0 & 0 & 0 & 0 & 0 & 0 & 0 & 0 & 0 & 0 & 266 & 1 & 28 & 3 & 25 & 1 & 0 & 0 & 0\\
119&0 & 0 & 1 & 0 & 0 & 0 & 0 & 0 & 0 & 0 & 0 & 0 & 0 & 0 & 0 & 264 & 1 & 28 & 3 & 24 & 1 & 0 & 0 & 0\\
120&0 & 0 & 1 & 0 & 0 & 0 & 0 & 0 & 0 & 0 & 0 & 0 & 0 & 0 & 0 & 261 & 1 & 29 & 3 & 21 & 1 & 0 & 0 & 0\\
121&0 & 0 & 1 & 0 & 0 & 0 & 0 & 0 & 0 & 0 & 0 & 0 & 0 & 0 & 0 & 268 & 1 & 28 & 2 & 20 & 1 & 0 & 0 & 0\\
122&0 & 0 & 1 & 0 & 0 & 0 & 0 & 0 & 0 & 0 & 0 & 0 & 0 & 0 & 0 & 274 & 1 & 28 & 3 & 20 & 1 & 2 & 0 & 0\\
123&0 & 0 & 1 & 0 & 0 & 0 & 0 & 0 & 0 & 0 & 0 & 0 & 0 & 0 & 0 & 276 & 1 & 28 & 2 & 19 & 1 & 9 & 0 & 0\\
124&0 & 0 & 1 & 0 & 0 & 0 & 0 & 0 & 0 & 0 & 0 & 0 & 0 & 0 & 0 & 277 & 1 & 27 & 2 & 19 & 0 & 29 & 0 & 0\\
125&0 & 0 & 1 & 0 & 0 & 0 & 0 & 0 & 0 & 0 & 0 & 0 & 0 & 0 & 0 & 279 & 1 & 26 & 2 & 18 & 0 & 48 & 0 & 0\\
126&0 & 0 & 1 & 0 & 0 & 0 & 0 & 0 & 0 & 0 & 0 & 0 & 0 & 0 & 0 & 277 & 1 & 24 & 2 & 15 & 0 & 61 & 0 & 0\\
127&0 & 0 & 1 & 0 & 0 & 0 & 0 & 0 & 0 & 0 & 0 & 0 & 0 & 0 & 0 & 274 & 1 & 24 & 2 & 13 & 0 & 73 & 0 & 0\\
128&0 & 0 & 1 & 0 & 0 & 0 & 0 & 0 & 0 & 0 & 0 & 0 & 0 & 0 & 0 & 275 & 0 & 24 & 2 & 13 & 0 & 79 & 0 & 0\\
129&0 & 0 & 1 & 0 & 0 & 0 & 0 & 0 & 0 & 0 & 0 & 0 & 0 & 0 & 0 & 282 & 0 & 24 & 1 & 12 & 0 & 92 & 1 & 0\\
130&0 & 0 & 1 & 0 & 0 & 0 & 0 & 0 & 0 & 0 & 0 & 0 & 0 & 0 & 0 & 278 & 0 & 24 & 1 & 12 & 0 & 103 & 5 & 0\\
131&0 & 0 & 1 & 0 & 0 & 0 & 0 & 0 & 0 & 0 & 0 & 0 & 0 & 0 & 0 & 275 & 0 & 24 & 0 & 11 & 0 & 109 & 17 & 0\\
132&0 & 0 & 1 & 0 & 0 & 0 & 0 & 0 & 0 & 0 & 0 & 0 & 0 & 0 & 0 & 261 & 0 & 24 & 0 & 11 & 0 & 112 & 37 & 0\\
133&0 & 0 & 1 & 0 & 0 & 0 & 0 & 0 & 0 & 0 & 0 & 0 & 0 & 0 & 0 & 225 & 0 & 22 & 0 & 10 & 0 & 113 & 110 & 0\\
134&0 & 0 & 1 & 0 & 0 & 0 & 0 & 0 & 0 & 0 & 0 & 0 & 0 & 0 & 0 & 182 & 0 & 22 & 0 & 10 & 0 & 114 & 176 & 0\\
135&0 & 0 & 1 & 0 & 0 & 0 & 0 & 0 & 0 & 0 & 0 & 0 & 0 & 0 & 0 & 159 & 0 & 22 & 0 & 10 & 0 & 114 & 209 & 0\\
136&0 & 0 & 1 & 0 & 0 & 0 & 0 & 0 & 0 & 0 & 0 & 0 & 0 & 0 & 0 & 127 & 0 & 22 & 0 & 7 & 0 & 112 & 247 & 0\\
137&0 & 0 & 1 & 0 & 0 & 0 & 0 & 0 & 0 & 0 & 0 & 0 & 0 & 0 & 0 & 105 & 0 & 23 & 0 & 6 & 0 & 109 & 287 & 2\\
138&0 & 0 & 1 & 0 & 0 & 0 & 0 & 0 & 0 & 0 & 0 & 0 & 0 & 0 & 0 & 96 & 0 & 23 & 0 & 6 & 0 & 111 & 299 & 14\\
139&0 & 0 & 1 & 0 & 0 & 0 & 0 & 0 & 0 & 0 & 0 & 0 & 0 & 0 & 0 & 89 & 0 & 23 & 0 & 6 & 0 & 117 & 310 & 45\\
140&0 & 0 & 1 & 0 & 0 & 0 & 0 & 0 & 0 & 0 & 0 & 0 & 0 & 0 & 0 & 80 & 0 & 22 & 0 & 4 & 0 & 115 & 319 & 51\\
141&0 & 0 & 1 & 0 & 0 & 0 & 0 & 0 & 0 & 0 & 0 & 0 & 0 & 0 & 0 & 67 & 0 & 23 & 0 & 4 & 0 & 118 & 335 & 36\\
142&0 & 0 & 1 & 0 & 0 & 0 & 0 & 0 & 0 & 0 & 0 & 0 & 0 & 0 & 0 & 65 & 0 & 23 & 0 & 3 & 0 & 118 & 342 & 19\\
143&0 & 0 & 1 & 0 & 0 & 0 & 0 & 0 & 0 & 0 & 0 & 0 & 0 & 0 & 0 & 55 & 0 & 22 & 0 & 3 & 0 & 116 & 352 & 6\\
144&0 & 0 & 1 & 0 & 0 & 0 & 0 & 0 & 0 & 0 & 0 & 0 & 0 & 0 & 0 & 52 & 0 & 22 & 0 & 3 & 0 & 109 & 359 & 2\\
145&0 & 0 & 1 & 0 & 0 & 0 & 0 & 0 & 0 & 0 & 0 & 0 & 0 & 0 & 0 & 51 & 0 & 23 & 0 & 3 & 0 & 101 & 363 & 2\\
146&0 & 0 & 1 & 0 & 0 & 0 & 0 & 0 & 0 & 0 & 0 & 0 & 0 & 0 & 0 & 50 & 0 & 24 & 0 & 3 & 0 & 93 & 364 & 7\\
147&0 & 0 & 1 & 0 & 0 & 0 & 0 & 0 & 0 & 0 & 0 & 0 & 0 & 0 & 0 & 50 & 0 & 27 & 0 & 3 & 0 & 93 & 369 & 7\\
148&0 & 0 & 1 & 0 & 0 & 0 & 0 & 0 & 0 & 0 & 0 & 0 & 0 & 0 & 0 & 45 & 0 & 26 & 0 & 3 & 0 & 97 & 378 & 8 \\
                          \bottomrule
                        \end{tabular}}
                    \end{center}
                    \end{table}

\section{Selection of 123 Solved Sequences}
Tables~\ref{tab:solvedseq1}, \ref{tab:solvedseq2}, \ref{tab:solvedseq3} present a sample of 123 sequences solved during the $\mathit{nmt}_0$ and $\mathit{nmt}_1$ runs. Their solutions found by the system are presented on our web page,\footnoteA{\url{https://github.com/Anon52MI4/oeis-alien}} both in our language and translated to Python.

\begin{table}
  \caption{Samples of the solved sequences.}\label{tab:solvedseq1}
          \begin{center}
            \resizebox{\textwidth}{!}{%
			\begin{tabular}{lp{14cm}}
                          \toprule
\url{https://oeis.org/A317485} &	       Number of Hamiltonian paths in the n-Bruhat graph.\\
\url{https://oeis.org/A349073} &	       a(n) = U(2*n, n), where U(n, x) is the Chebyshev polynomial of the second kind.\\
\url{https://oeis.org/A293339} &	       Greatest integer k such that $k/2^n < 1/e$.\\
\url{https://oeis.org/A1848} &	       Crystal ball sequence for 6-dimensional cubic lattice.\\
\url{https://oeis.org/A8628} &	       Molien series for $A_5$.\\
\url{https://oeis.org/A259445} &	       Multiplicative with $a(n) = n$ if n is odd and $a(2^s)=2$.\\
\url{https://oeis.org/A314106} &	       Coordination sequence Gal.6.199.4 where G.u.t.v denotes the coordination sequence for a vertex of type v in tiling number t in the Galebach list of u-uniform tilings\\
\url{https://oeis.org/A311889} &	       Coordination sequence Gal.6.129.2 where G.u.t.v denotes the coordination sequence for a vertex of type v in tiling number t in the Galebach list of u-uniform tilings.\\
\url{https://oeis.org/A315334} &	       Coordination sequence Gal.6.623.2 where G.u.t.v denotes the coordination sequence for a vertex of type v in tiling number t in the Galebach list of u-uniform tilings.\\
\url{https://oeis.org/A315742} &	       Coordination sequence Gal.5.302.5 where G.u.t.v denotes the coordination sequence for a vertex of type v in tiling number t in the Galebach list of u-uniform tilings.\\
\url{https://oeis.org/A004165} &	       OEIS writing backward\\
\url{https://oeis.org/A83186} &	       Sum of first n primes whose indices are primes.\\
\url{https://oeis.org/A88176} &	       Primes such that the previous two primes are a twin prime pair.\\
\url{https://oeis.org/A96282} &	       Sums of successive twin primes of order 2.\\
\url{https://oeis.org/A53176} &	       Primes p such that $2p+1$ is composite.\\
\url{https://oeis.org/A267262} &	       Total number of OFF (white) cells after n iterations of the "Rule 111" elementary cellular automaton starting with a single ON (black) cell.\\
\url{https://oeis.org/A273385} &	       Number of active (ON,black) cells at stage $2^n-1$ of the two-dimensional cellular automaton defined by "Rule 659", based on the 5-celled von Neumann neighborhood.\\
\url{https://oeis.org/A60431} &	       Number of cubefree numbers <= n.\\
\url{https://oeis.org/A42731} &	       Denominators of continued fraction convergents to sqrt(895).\\
\url{https://oeis.org/A81495} &	       Start with Pascal's triangle; form a rhombus by sliding down n steps from top on both sides then sliding down inwards to complete the rhombus and then deleting the inner numbers; a(n) = sum of entries on perimeter of rhombus.\\
\url{https://oeis.org/A20027} &	       Nearest integer to Gamma(n + 3/8)/Gamma(3/8).\\
\url{https://oeis.org/A99197} &	       Figurate numbers based on the 10-dimensional regular convex polytope called the 10-dimensional cross-polytope, or 10-dimensional hyperoctahedron, which is represented by the Schlaefli symbol {3, 3, 3, 3, 3, 3, 3, 3, 4}. It is the dual of the 10-dimensional hypercube.\\
\url{https://oeis.org/A220469} &	       Fibonacci 14-step numbers, $a(n) = a(n-1) + a(n-2) + ... + a(n-14)$.\\
\url{https://oeis.org/A8583} &	       Molien series for Weyl group $E_7$.\\
\url{https://oeis.org/A251672} &	       8-step Fibonacci sequence starting with 0,0,0,0,0,0,1,0.\\
\url{https://oeis.org/A124615} &	       Poincaré series [or Poincare series] $P(T_{3,2}; x)$.\\
\url{https://oeis.org/A79262} &	       Octanacci numbers: $a(0)=a(1)=...=a(6)=0, a(7)=1;$ for $n >= 8, a(n) = Sum_{i=1..8} a(n-i)$.\\
\url{https://oeis.org/A75068} &	       Product of prime(n) primes starting from prime(n).\\
\url{https://oeis.org/A57168} &	       Next larger integer with same binary weight (number of 1 bits) as n.\\
\url{https://oeis.org/A1553} &	       $a(n) = 1^n + 2^n + ... + 6^n$.\\
\url{https://oeis.org/A19560} &	       Coordination sequence for $C_4$ lattice.\\
\url{https://oeis.org/A289834} &	       Number of perfect matchings on n edges which represent RNA secondary folding structures characterized by the Lyngso and Pedersen (L\&P) family and the Cao and Chen (C\&C) family.\\
\url{https://oeis.org/A5249} &	       Determinant of inverse Hilbert matrix.\\
\url{https://oeis.org/A3714} &	       Fibbinary numbers: if $n = F(i1) + F(i2) + ... + F(ik)$ is the Zeckendorf representation of n (i.e., write n in Fibonacci number system) then $a(n) = 2^{i1 - 2} + 2^{i2 - 2} + ... + 2^{ik - 2}$. Also numbers whose binary representation contains no two adjacent 1's.\\
\url{https://oeis.org/A4457} &	       Nimsum $n + 16$.\\
\url{https://oeis.org/A92143} &	       Cumulative product of all divisors of 1..n.\\
\url{https://oeis.org/A2119} &	       Bessel polynomial $y_n(-2)$.\\
\url{https://oeis.org/A5913} &	       $a(n) = [ tau * a(n-1) ] + [ tau * a(n-2) ]$.\\
\url{https://oeis.org/A34960} &	       Divide odd numbers into groups with prime(n) elements and add together.\\
\url{https://oeis.org/A247395} &	       The smallest numbers of every class in a classification of positive numbers (see comment).\\
\url{https://oeis.org/A68068} &	       Number of odd unitary divisors of n. d is a unitary divisor of n if d divides n and $gcd(d,n/d)=1$. \\ 
\bottomrule
                        \end{tabular}}
                    \end{center}
                  \end{table}

\begin{table}
  \caption{Samples of the solved sequences.}\label{tab:solvedseq2}
          \begin{center}
            \resizebox{\textwidth}{!}{%
			\begin{tabular}{lp{13cm}}
                          \toprule
\url{https://oeis.org/A30973} &	       \  [ $exp(1/5) * n!$ ].\\
\url{https://oeis.org/A54469} &	       A second-order recursive sequence.\\
\url{https://oeis.org/A54054} &	       Smallest digit of n.\\
\url{https://oeis.org/A36561} &	       Nicomachus triangle read by rows, $T(n, k) = 2^{n - k} * 3^k, for 0 <= k <= n$.\\
\url{https://oeis.org/A107347} &	       Number of even semiprimes strictly between prime(n) and 2 * prime(n).\\
\url{https://oeis.org/A123379} &	       Values x of the solutions (x,y) of the Diophantine equation $5*(X-Y)^4 - 4XY = 0$ with X >= Y.\\
\url{https://oeis.org/A201204} &	       Half-convolution of Catalan sequence A000108 with itself.\\
\url{https://oeis.org/A125494} &	       Composite evil numbers.\\
\url{https://oeis.org/A277094} &	       Numbers k such that $sin(k) > 0$ and $sin(k+2) < 0$.\\
\url{https://oeis.org/A59760} &	       a(n) is the number of edges (one-dimensional faces) in the convex polytope of real n X n doubly stochastic matrices.\\
\url{https://oeis.org/A246303} &	       Numbers k such that $cos(k) < cos(k+1)$.\\
\url{https://oeis.org/A7957} &	       Numbers that contain an odd digit.\\
\url{https://oeis.org/A7452} &	       Expand $cos x / exp x$ and invert nonzero coefficients.\\
\url{https://oeis.org/A88896} &	       Length of longest integral ladder that can be moved horizontally around the right angled corner where two hallway corridors of integral widths meet.\\
\url{https://oeis.org/A131989} &	       Start with the symbol *| and for each iteration replace * with *| . This sequence is the number of *'s between each dash.\\
\url{https://oeis.org/A308066} &	       Number of triangles with perimeter n whose side lengths are even.\\
\url{https://oeis.org/A11540} &	       Numbers that contain a digit 0.\\
\url{https://oeis.org/A156660} &	       Characteristic function of Sophie Germain primes.\\
\url{https://oeis.org/A167132} &	       Gaps between twin prime pairs.\\
\url{https://oeis.org/A8846} &	       Hypotenuses of primitive Pythagorean triangles.\\
\url{https://oeis.org/A332381} &	       a(n) is the Y-coordinate of the n-th point of the Peano curve. Sequence A332380 gives X-coordinates.\\
\url{https://oeis.org/A25492} &	       Fixed point reached by iterating the Kempner function A002034 starting at n.\\
\url{https://oeis.org/A131530} &	       Numbers k such that $k^2 - k - 1$ and $k^2 - k + 1$ are twin primes.\\
\url{https://oeis.org/A143165} &	       Expansion of the exponential generating function $arcsin(2x)/(2(1-2*x)^{3/2})$.\\
\url{https://oeis.org/A30957} &	       \ [ $exp(1/9)*n!$ ].\\
\url{https://oeis.org/A295286} &	       Sum of the products of the smaller and larger parts of the partitions of n into two parts with the smaller part odd.\\
\url{https://oeis.org/A86699} &	       Number of n X n matrices over GF(2) with rank n-1.\\
\url{https://oeis.org/A2819} &	       Liouville's function L(n) = partial sums of A008836.\\
\url{https://oeis.org/A7318} &	       Pascal's triangle read by rows: $C(n,k) = binomial(n,k) = n!/(k!*(n-k)!), 0 <= k <= n$.\\
\url{https://oeis.org/A8836} &	       Liouville's function lambda(n) =$ (-1)^k$, where k is number of primes dividing n (counted with multiplicity).\\
\url{https://oeis.org/A266776} &	       Molien series for invariants of finite Coxeter group $A_7$.\\
\url{https://oeis.org/A284115} &	       Hosoya triangle of Lucas type.\\
\url{https://oeis.org/A45717} &	       For each prime p take the sum of nonprimes < p.\\
\url{https://oeis.org/A307508} &	       Primes p for which the continued fraction expansion of sqrt(p) does not have a 1 in the second position.\\
\url{https://oeis.org/A3506} &	       Triangle of denominators in Leibniz's Harmonic Triangle $a(n,k), n >= 1, 1 <= k <= n$.\\
\url{https://oeis.org/A93017} &	       Luhn algorithm double-and-add sum of digits of n.\\
\url{https://oeis.org/A121373} &	       Expansion of$ f(x) = f(x, -x^2)$ in powers of x where f(, ) is Ramanujan's general theta function.\\
\url{https://oeis.org/A227127} &	       The Akiyama-Tanigawa algorithm applied to 1/(1,2,3,5,... old prime numbers). Reduced numerators of the second row.\\
\url{https://oeis.org/A39637} &	       Number of steps to fixed point of "$n -> n/2 or (n+1)/2$ until result is prime".\\
\url{https://oeis.org/A548} &	       Squares that are not the sum of 2 nonzero squares.\\
\url{https://oeis.org/A131650} &	       Number of symbols in Babylonian numeral representation of n.\\
\bottomrule
                        \end{tabular}}
                    \end{center}
                  \end{table}

\begin{table}
  \caption{Samples of the solved sequences.}\label{tab:solvedseq3}
          \begin{center}
\resizebox{\textwidth}{!}{%
			\begin{tabular}{lp{13cm}}
                          \toprule
\url{https://oeis.org/A3538} &	       Divisors of $2^{30} - $1.\\
\url{https://oeis.org/A152135} &	       Maximal length of rook tour on an n X n+4 board.\\
\url{https://oeis.org/A8637} &	       Number of partitions of n into at most 8 parts.\\
\url{https://oeis.org/A113953} &	       A Jacobsthal triangle.\\
\url{https://oeis.org/A3605} &	       Unique monotonic sequence of nonnegative integers satisfying $a(a(n)) = 3n$.\\
\url{https://oeis.org/A266214} &	       Numbers n that are not coprime to the numerator of $zeta(2n)/(Pi^{2n})$.\\
\url{https://oeis.org/A266778} &	       Molien series for invariants of finite Coxeter group $A_9$.\\
\url{https://oeis.org/A101608} &	       Solution to Tower of Hanoi puzzle encoded in pairs with the moves:\\
& $(1,2),(2,3),(3,1),(2,1),(3,2),(1,3)$.\\
&The disks are moved from peg 1 to 2. For a tower of k disks use the first $2^k-1$ number pairs.\\
\url{https://oeis.org/A90971} &	       Sierpiński's triangle, read by rows, starting from 1:\\ 
                              &       $T(n,k) = (T(n-1,k) + T(n-1,k-1))\ \mmod\ 2$.\\
\url{https://oeis.org/A83743} &	       $a(1) = 1$; if $a(n-1) + n$ is prime then $a(n) = a(n-1) + n$, else $a(n) = a(n-1)$.\\
\url{https://oeis.org/A79683} &	       Order of Burnside group $B(6,n)$ of exponent 6 and rank n.\\
\url{https://oeis.org/A34444} &	       a(n) is the number of unitary divisors of n (d such that d divides n, $gcd(d, n/d) = 1)$.\\
\url{https://oeis.org/A218509} &	       Number of partitions of n in which any two parts differ by at most 7.\\
\url{https://oeis.org/A65109} &	       Triangle $T(n,k)$ of coefficients relating to Bezier curve continuity.\\
\url{https://oeis.org/A166555} &	       Triangle read by rows, Sierpinski's gasket, A047999 * (1,2,4,8,...) diagonalized.\\
\url{https://oeis.org/A119467} &	       A masked Pascal triangle.\\
\url{https://oeis.org/A194887} &	       Numbers that are the sum of two powers of 12.\\
\url{https://oeis.org/A1824} &	       Central factorial numbers.\\
\url{https://oeis.org/A47780} &	       Number of inequivalent ways to color faces of a cube using at most n colors.\\
\url{https://oeis.org/A8640} &	       Number of partitions of n into at most 11 parts.\\
\url{https://oeis.org/A29635} &	       The (1,2)-Pascal triangle (or Lucas triangle) read by rows.\\
\url{https://oeis.org/A45995} &	       Rows of Fibonacci-Pascal triangle.\\
\url{https://oeis.org/A5045} &	       Number of restricted 3 X 3 matrices with row and column sums n.\\
\url{https://oeis.org/A971} &	       Fermat coefficients.\\
\url{https://oeis.org/A5835} &	       Pseudoperfect (or semiperfect) numbers n: some subset of the proper divisors of n sums to n.\\
\url{https://oeis.org/A266773} &	       Molien series for invariants of finite Coxeter group $D_{10}$ (bisected).\\
\url{https://oeis.org/A69209} &	       Orders of non-Abelian Z-groups.\\
\url{https://oeis.org/A8383} &	       Coordination sequence for $A_4$ lattice.\\
\url{https://oeis.org/A70896} &	       Determinant of the Cayley addition table of $Z_{n}$.\\
\url{https://oeis.org/A262} &	       Number of "sets of lists": number of partitions of {1,...,n} into any number of lists, where a list means an ordered subset.\\
\url{https://oeis.org/A23436} &	       Dying rabbits: $a(n) = a(n-1) + a(n-2) - a(n-6)$.\\
\url{https://oeis.org/A8641} &	       Number of partitions of n into at most 12 parts.\\
\url{https://oeis.org/A68764} &	       Generalized Catalan numbers.\\
\url{https://oeis.org/A7856} &	       Subtrees in rooted plane trees on n nodes.\\
\url{https://oeis.org/A271} &	       Sums of ménage numbers.\\
\url{https://oeis.org/A199033} &	       Number of ways to place n non-attacking bishops on a 2 X 2n board.\\
\url{https://oeis.org/A1006} &	       Motzkin numbers: number of ways of drawing any number of nonintersecting chords joining n (labeled) points on a circle.\\
\url{https://oeis.org/A239768} &	       Number of pairs of functions (f,g) from a set of n elements into itself satisfying $f(x) = f(g(f(x)))$.\\
\url{https://oeis.org/A2895} &	       Domb numbers: number of 2n-step polygons on diamond lattice.\\
\url{https://oeis.org/A14342} &	       Convolution of primes with themselves.\\
\url{https://oeis.org/A27847} &          $a(n) = Sum_{d|n} sigma(n/d)*d^3$.\\
\bottomrule
                       \end{tabular}}
                    \end{center}
                  \end{table}

                \end{document}